\documentclass[10pt,twocolumn,letterpaper]{article}

\usepackage[pagenumbers]{cvpr} % To force page numbers, e.g. for an arXiv version

\usepackage{graphicx}
\usepackage{amsmath}
\usepackage{amssymb}
\usepackage{booktabs}
\usepackage{multirow}
\usepackage[dvipsnames]{xcolor}

\usepackage[pagebackref,breaklinks,colorlinks]{hyperref}

\usepackage[capitalize]{cleveref}
\crefname{section}{Sec.}{Secs.}
\Crefname{section}{Section}{Sections}
\Crefname{table}{Table}{Tables}
\crefname{table}{Tab.}{Tabs.}

\newcommand\blfootnote[1]{%
  \begingroup
  \renewcommand\thefootnote{}\footnote{#1}%
  \addtocounter{footnote}{-1}%
  \endgroup
}

\def\cvprPaperID{2} % *** Enter the CVPR Paper ID here
\def\confName{CVSports}
\def\confYear{2023}
\newcommand{\mysection}[1]{\vspace{2pt}\noindent\textbf{#1}}
\usepackage{xcolor, color, colortbl}         % colors
\definecolor{Highlight}{HTML}{39b54a}  % green

\usepackage{amssymb}% http://ctan.org/pkg/amssymb
\usepackage{pifont}% http://ctan.org/pkg/pifont
\usepackage{algorithm}
\usepackage{listings}
\usepackage{etoolbox}
\makeatletter
\AfterEndEnvironment{algorithm}{\let\@algcomment\relax}
\AtEndEnvironment{algorithm}{\kern2pt\hrule\relax\vskip3pt\@algcomment}
\let\@algcomment\relax
\newcommand\algcomment[1]{\def\@algcomment{\footnotesize#1}}
\renewcommand\fs@ruled{\def\@fs@cfont{\bfseries}\let\@fs@capt\floatc@ruled
  \def\@fs@pre{\hrule height.8pt depth0pt \kern2pt}%
  \def\@fs@post{}%
  \def\@fs@mid{\kern2pt\hrule\kern2pt}%
  \let\@fs@iftopcapt\iftrue}
\makeatother
\newcommand{\cmmnt}[1]{}

\usepackage[normalem]{ulem}

\usepackage{listings}
\usepackage{color}
\definecolor{codegreen}{rgb}{0,0.6,0}
\definecolor{codegray}{rgb}{0.5,0.5,0.5}
\definecolor{codepurple}{rgb}{0.58,0,0.82}
\definecolor{backcolour}{rgb}{1,1,1}
 
\lstdefinestyle{mystyle}{
    backgroundcolor=\color{backcolour},   
    commentstyle=\color{codegreen},
    keywordstyle=\color{magenta},
    numberstyle=\tiny\color{codegray},
    stringstyle=\color{codepurple},
    basicstyle=\footnotesize,
    breakatwhitespace=false,         
    breaklines=true,                 
    captionpos=b,                    
    keepspaces=true,                 
    numbers=left,                    
    numbersep=5pt,                  
    showspaces=false,                
    showstringspaces=false,
    showtabs=false,                  
    tabsize=2
}
 
\makeatletter

\newcommand{\Rmnum}[1]{\expandafter\@slowromancap\romannumeral #1@}
\makeatother
\begin{document}

% \title{MVAS: Multi-View Soccer Videos Understanding for Referees Assistance}
% \title{MVAS: a Multi-View Assistant System to Assist Referee in Fouls Decision Making}
% \title{Leveling Up Fair Play in Soccer: A Multi-View Assistant System for Accurate Referee Decision Making on Fouls}
% \title{MVAS: Multi-View Assistant System \\for Accurate Referee Decision Making in Soccer}
\title{VARS: Video Assistant Referee System\\for Automated Soccer Decision Making from Multiple Views}
% \title{VARS: Towards Automated Soccer Decision Making from Multiple Views}

\author{Jan Held$^{*,1}$%\\ {\small Université Libre de Bruxelles} \and
\quad
Anthony Cioppa$^{*,1,2}$  %\\ {\small University of Liège, KAUST}  \and
\quad
Silvio Giancola$^{*,2}$  %\\ {\small KAUST}  \and 
\and
Abdullah Hamdi$^{2}$
\quad
Bernard Ghanem$^{2}$  %\\ {\small KAUST}  \and
\quad
Marc Van Droogenbroeck$^{1}$  %\\ {\small University of Liège}
\and $^1$ {\small University of Liège}
\quad $^2$ {\small KAUST}
}
\maketitle

%%%%%%%%% ABSTRACT
\begin{abstract}
% GENERIC INTRO OF THE FIELD
The Video Assistant Referee (VAR) has revolutionized association football, enabling referees to review incidents on the pitch, make informed decisions, and ensure fairness.
% CURRENT ISSUE IN THAT FIELD
However, due to the lack of referees in many countries and the high cost of the VAR infrastructure, only professional leagues can benefit from it.
% OUR SOLUTION
In this paper, we propose a \emph{Video Assistant Referee System (VARS)} that can automate soccer decision-making.
% MORE INFO ON THAT SOLUTION
VARS leverages the latest findings in multi-view video analysis, to provide real-time feedback to the referee, and help them make informed decisions that can impact the outcome of a game.
% MORE ON EXPERIMENTAL VALIDATION
To validate VARS, we introduce \emph{SoccerNet-MVFoul}, a novel video dataset of soccer fouls from multiple camera views, annotated with extensive foul descriptions by a professional soccer referee, and we benchmark our VARS to automatically recognize the characteristics of these fouls.
% To validate VARS, we introduce a novel dataset, called SoccerNet-MVFoul, which extracts relevant fouls in soccer broadcasts from multiple views.
% We curate this data with extensive properties, such as if the action is an offence, the foul type, and its severity. A professional soccer referee manually annotated all properties without seeing the decision of the on-field referee to avoid any bias toward their decision.
% GENERIC CONCLUSION ON THE IMPACT
% By combining the latest computer vision techniques with the needs of referees, 
We believe that VARS has the potential to revolutionize soccer refereeing and take the game to new heights of fairness and accuracy across all levels of professional and amateur federations. %Our dataset and code will be released upon publication.
\blfootnote{\textbf{(*)} Denotes equal contributions \\Contacts: jan.held@student.uliege.be, silvio.giancola@kaust.edu.sa, anthony.cioppa@uliege.be. 
Data/code available at \href{https://www.soccer-net.org}{www.soccer-net.org}.
}
\end{abstract}

% GENERIC INTRO OF THE FIELD
% The introduction of Video Assistant Referees (VARs) in association football has been nothing short of a revolution, transforming the game and the way it is played. 
% With VARs, referees can review incidents on the pitch and make informed decisions, avoiding mistakes and ensuring fairness. 

% CURRENT ISSUE IN THAT FIELD
% The introduction of this technology has sparked intense debate and controversy, with some arguing that it disrupts the flow of the game, while others believe that it is a crucial tool for ensuring accuracy and justice. 
% Despite the differing opinions, there is no doubt that VARs have made a significant impact on football, leading to a renewed focus on the role of technology in sports. 

\begin{figure}[t]
    \centering
    \includegraphics[width=\linewidth]{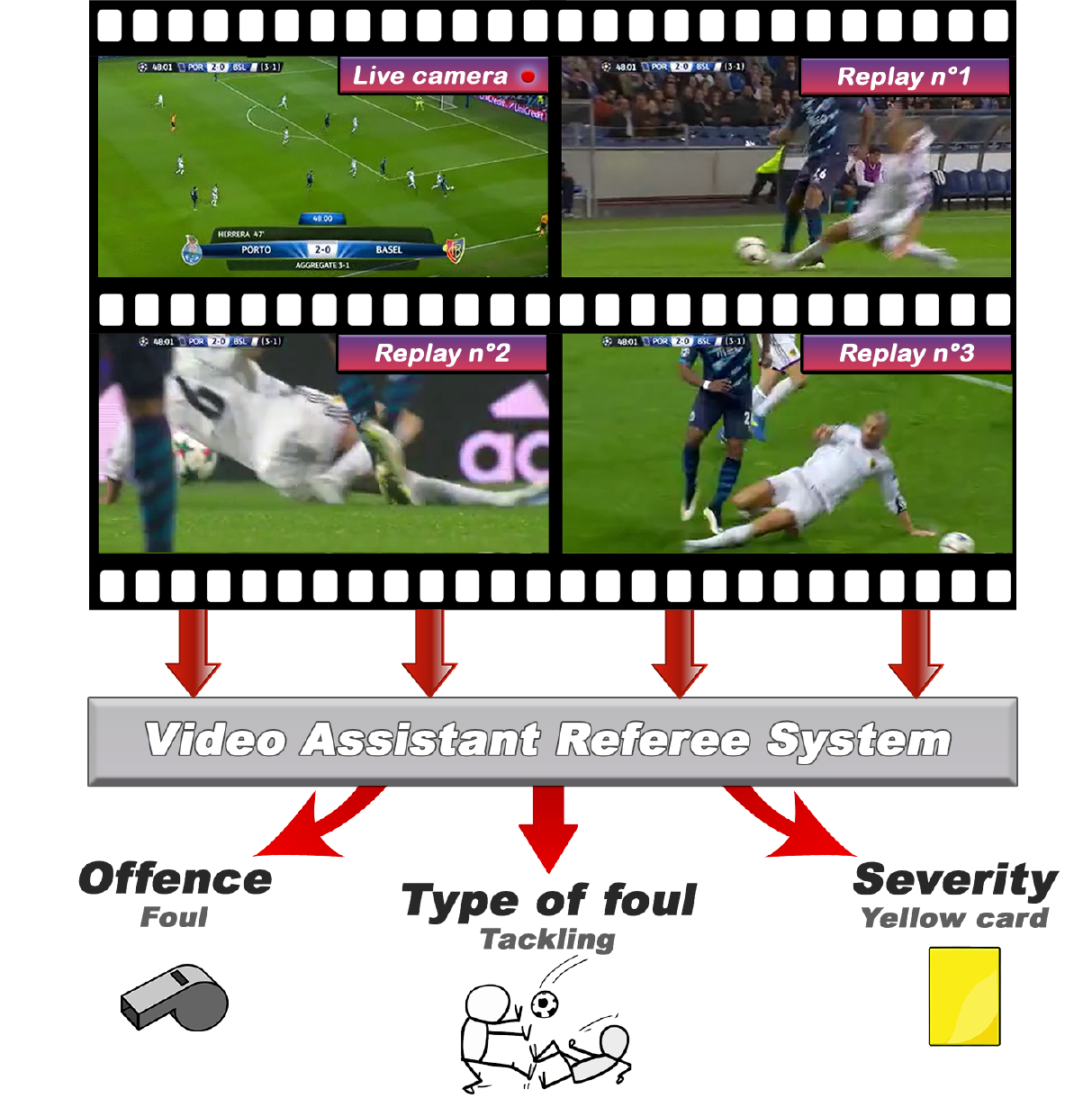}
    \caption{\textbf{Video Assistant Referee System (VARS).} We propose an automated VARS for automatically classifying whether an action is a foul, determining the type of foul (\eg, `Tackling', `Pushing', etc.), and the appropriate punishment the player should receive for the foul (\ie, `No card', `Yellow card', or `Red card') from a multi-view camera setup.}
    \label{fig:graphical_abstract}
\end{figure}

%%%%%%%%% BODY TEXT
\section{Introduction}
\label{sec:intro}

Over the past decades, the technology used by referees in soccer has undergone a drastic evolution. 
Before the beginning of this century, referees and their assistants only relied on their own judgment, and the communication between them was based on eye contact and body language. 
The French and Scottish refereeing trios were the first ones to be linked through wireless mini-earphones during league matches, facilitating communication among them~\cite{UEFA2003Headsets}. 
Nowadays, wireless headsets are essential pieces of equipment for referees worldwide, made mandatory for high-level competitions.
Another important breakthrough in professional soccer was the introduction of goal-line technology, which uses a combination of cameras and sensors to determine whether the entire ball has crossed the goal line or not. 
This technology aims to prevent controversial goals such as the famous ``ghost goal'' scored by Geoff Hursts in the 1966 World Cup final against Germany, where the ball may not have fully crossed the line and led to England receiving the world champion title~\cite{FIFA2022GoalLineTechnology}. 
Successively, the International Football Association Board (IFAB) approved the introduction of extra referees, namely the video assistant referees (VAR), to prevent game-changing errors. 
% Successively, the International Football Association Board (IFAB) approved the video assistant referee (VAR) to assist referees in the event of game-changing decision making. 
% More recently, the International Football Association Board (IFAB) approved the use of the video assistant referee (VAR) in football to assist referees in the event of game-changing decision making. 
More recently, artificial intelligence systems appeared for the first time during the World Cup 2022 in Qatar.
Semi-automated offside technology now supports the VAR to help referees make faster, more accurate, and more reproducible offside decisions~\cite{FIFA2023SemiAutomatedOffsideTechnology}. 
This new system relies on $12$ well-calibrated cameras to track the ball and the player's body pose, sending an automated offside alert to the video assistant referee inside the video operation room. 
% This new system uses $12$ well-calibrated cameras mounted underneath the roof of the stadium to track the ball and up to $29$ data points of each individual player calculating their exact pose and position on the pitch. 
% By analyzing the body and ball tracking data, the new technology provides an automated offside alert to the video assistant referee inside the video operation room~\cite{FIFA2023SemiAutomatedOffsideTechnology}. 
This shows that soccer is moving towards more assistance or even automated systems to help referees make better decisions.

However, despite its intention to improve the accuracy of referee decisions, VAR has become a source of frustration and anger for many football fans around the world. 
Since we have a different video assistant referee for each game, we do not always have consistent decisions. 
Sometimes the VAR predicts different outcomes for similar situations in different games and leagues. 
%This inconsistency can lead to confusion and frustration among players, coaches, and fans.
Moreover, the implementation of the VAR technology and infrastructure requires a substantial financial investment, limiting its accessibility to only the top-tier leagues and clubs. 
As a result, semi-professional or amateur leagues are unable to benefit from the VAR due to financial constraints. Additionally, the shortage of referees worldwide makes it impossible in staffing additional referees as Video Assistant Referees, except for the professional leagues. 

In this work, we propose a first step towards a fully automated ``Video Assistant Referee System'' (VARS) which could support or replace the current VAR.
We attempt to automatically predict all fouls and suggest appropriate sanctions to the players.
In case the on-field referee makes a significant mistake, our VARS could intervene to suggest a revision. 
It is intended that, just like regular VAR, our VARS serves as a support system for the referee, but the final decision remains in the hands of the on-field referee.
To achieve this objective, we rely on multi-view uncalibrated camera video streams, which are already leveraged to edit broadcast games.
Specifically, we release a new dataset comprising $3{,}901$ actions with multi-view clips of $5$ seconds around the action, annotated by a professional referee. 
We focus our analysis on the classification of foul types and evaluate their severity to identify the sanction for the player.
% sanction the player with a yellow or red card. 
%
Practically, our VARS analyses the different streams and combines the information from the multiple cameras. We show that using a multi-view system largely improves the performance compared to a single view and that we reach good performance on our video recognition tasks.

\mysection{Contributions.} We summarize our contribution as follows:
\textbf{(i)} We publicly release \textit{SoccerNet-MVFouls}, a new multi-view video dataset containing video clips captured by multiple cameras, annotated with $10$ properties. % and two new video recognition tasks.
\textbf{(ii)} We propose \textit{VARS}, a new multi-camera video recognition system for classifying fouls and their severity.
\textbf{(iii)} We propose a thorough study of using multiple views and how different types of camera views can influence the performance of VARS on two new video recognition tasks.

\section{Related work}
\label{sec:related_work}

\begin{table*}[t]
    \centering
    \resizebox{\linewidth}{!}{% <------ Don't forget this %
    \begin{tabular}{l|ccccc}
    \toprule
\textbf{Dataset} & \textbf{Context} &\textbf{Task} & \textbf{Videos/Images} & \textbf{View} & \textbf{Type}\\\midrule
Kinetics 400~\cite{Kay2017TheKinetics-arxiv}   &  Human actions &Classification &   $650{,}000$  & Single-view  & Videos  \\ \rowcolor[HTML]{EFEFEF} 
NTU RGB+D 120~\cite{Liu2020NTURGBD}   & Human actions &Classification &   $114{,}480$  & Multi-view & Videos   \\
 Northwestern-UCLA Multiview~\cite{wang2014cross}  & Human actions &Classification &$1{,}493$  & Multi-view & Videos \\ \rowcolor[HTML]{EFEFEF} 
UWA3D Multiview II~\cite{rahmani2016histogram}  & Human actions &Classification  &$900$  & Multi-view & Videos  \\
SoccerNet-v2 (Actions)~\cite{Giancola2022SoccerNet}  & Soccer &Classification &    $110{,}458$  & Single-View & Videos \\ \rowcolor[HTML]{EFEFEF} 
SoccerNet-v3 (Re-id.) ~\cite{Cioppa2022Scaling}  &   Soccer & 
Re-identifcation  &   $33{,}986$  & Multi-View & Images\\ \midrule
\textbf{SoccerNet-MVFouls (Ours)} & Soccer (Fouls) &Classification & $8{,}923$  & Multi-view & Videos \\ \bottomrule
    \end{tabular}
    }
    \caption{ \textbf{Video action understanding datasets}. Comparative overview of relevant datasets for multi/single-view action recognition in videos/images. Our dataset is the only one providing multi-view videos for classification in sports, with 10 annotated properties per action.}
    \label{tab:MOT_Comparison}
\end{table*}

\mysection{Sports understanding.}
As a research topic, sports video understanding has increased in popularity thanks to its challenging and fine-grained nature~\cite{Moeslund2014Computer,Thomas2017Computer}.
Nowadays, most state-of-the-art automatic methods are based on deep learning and have shown impressive performance on tasks such as player detection and tracking~\cite{Cioppa2020Multimodal, Maglo2022Efficient, Vandeghen2022SemiSupervised}, tactics analysis~\cite{Suzuki2019Team}, pass feasibility~\cite{ArbuesSanguesa2020Using} and prediction in soccer~\cite{Honda2022Pass}, talent scouting~\cite{Decroos2019Actions}, or player re-identification in occluded scenarios~\cite{Somers2023Body}.
Video classification started as a key area of research in this field~\cite{Wu2022ASurvey-arxiv}, with approaches proposed to recognize specific actions~\cite{Saraogi2016Event,Khan2018Learning} or distinguish between different game phases~\cite{Cioppa2018ABottomUp}. 
With the growing interest in temporal activity localization~\cite{Caba2015ActivityNet}, the task of action spotting~\cite{Cartas2022AGraphbased,Cioppa2020AContextAware, Darwish2022STE,Hong2022Spotting,Soares2022Action-arxiv, Soares2022Temporally,Zhu2022ATransformerbased} has gained interest as it provides precise localization of specific actions within a soccer game. 
% Action spotting enables the automatic generation of highlights for games' most salient moments and increases fan engagement~\cite{Cioppa2020AContextAware, Gao2020Automatic}.

% The progress in those tasks was made possible thanks to the availability of large-scale datasets such as the ones of Pappalardo~\etal~\cite{Pappalardo2019Apublic}, Yu~\etal~\cite{Yu2018Comprehensive}, SoccerTrack~\cite{Scott2022SoccerTrack}, SoccerDB~\cite{Jiang2020SoccerDB}, and DeepSportRadar~\cite{VanZandycke2022DeepSportradarv1}.
% Giancola~\etal~\cite{Giancola2018SoccerNet} introduced the SoccerNet dataset, which has grown to be the most extensive collection of data and annotations for video understanding in soccer, including  benchmarks for $10$ different tasks, such as action spotting~\cite{Deliege2021SoccerNetv2}, camera calibration~\cite{Cioppa2022Scaling}, and player tracking~\cite{Cioppa2022SoccerNetTracking}. 
% Cioppa~\etal~\cite{Cioppa2022Scaling} also introduce the task of player re-identification across multiple views by simulating a multi-camera system from a broadcast feed using the replays. 

The progress in those tasks was made possible thanks to the availability of large-scale datasets~\cite{Pappalardo2019Apublic,Yu2018Comprehensive,Scott2022SoccerTrack,Jiang2020SoccerDB,VanZandycke2022DeepSportradarv1}.
Giancola~\etal~\cite{Giancola2018SoccerNet} introduced the SoccerNet dataset, which has grown to be the most extensive collection of data and annotations for video understanding in soccer, including benchmarks for $10$ different tasks, ranging from broadcast understanding~\cite{Deliege2021SoccerNetv2}, field understanding~\cite{Cioppa2022Scaling} and player understanding~\cite{Cioppa2022SoccerNetTracking}. 
The SoccerNet team also organizes yearly competitions on these different tasks to foster research in the field~\cite{Giancola2022SoccerNet}. 
The dataset presented in this paper extends SoccerNet by proposing a novel multi-view video collection including foul annotations for video recognition tasks.

\mysection{Video understanding.}
For a long time, video understanding lagged behind image understanding due to the lack of large-scale video datasets such as ImageNet or CIFAR-100~\cite{deng2009imagenet,krizhevsky2009learning} in the video domain. 
However, the release of large video understanding datasets such as UCF101~\cite{Soomro2012UCF101-arxiv}, ActivityNet~\cite{Caba2015ActivityNet}, YouTube-8M~\cite{AbuElHaija2016YouTube8M-arxiv}, and Kinetics~\cite{Kay2017TheKinetics-arxiv} has led to a surge in popularity and interest in the field. 
Video understanding tasks include video classification~\cite{karpathy2014large, yue2015beyond, feichtenhofer2019slowfast}, action recognition ~\cite{Simonyan2014Two, wang2016temporal}, video captioning~\cite{krishna2017dense, gao2017video, wang2018video}, and video generation~\cite{li2018video}.

% In video classification, there has been a growing interest in developing video classification models that can capture spatio-temporal features effectively.
The interest in developing video classification models that capture spatio-temporal information has significantly grown. 
%
% Temporal Segment Network
Temporal Segment Network (TSN)~\cite{wang2018temporal} aggregates features across multiple temporal video segments to improve recognition performance.
%
% R2plus1d
Tran~\etal~\cite{tran2018closer} proposed a new spatio-temporal convolutional block R(2+1)D and analyze its effect on action recognition models.
%
% Multiscale vision transformer 
Recently, the Multiscale Vision Transformer (MViT)~\cite{li2022mvitv2, fan2021multiscale} came as a way to combine the strengths of both convolutional neural networks (CNNs) and transformers for video classification, capturing both spatial and temporal attentions. 
% Recently, there has been increasing interest in combining the strengths of both convolutional neural networks (CNNs) and transformers for video classification. 
% The Multiscale Vision Transformer (MViT)~\cite{li2022mvitv2, fan2021multiscale} is one such method that integrates a multiscale feature representation with a transformer-based architecture to capture both spatial and temporal information in videos.
In this work, we train different video representations to learn per-clip features that we aggregate from multiple views to identify the different properties of the fouls.

% info the clips from multiple view representing fouls  and identify the 
% The MViT model achieves state-of-the-art performance on several benchmark video classification datasets.
% In this work, our approach involves a multi-view configuration. 
% Instead of relying on a single video input, we extract spatial and temporal features from various viewpoints and we combine them to obtain a final prediction.

\mysection{Multi-view understanding.}
Su~\etal~\cite{mvcnn} introduces the idea of training image encoders to recognize 3D objects from multiple views, benefiting from the mature 2D computer vision. 
Most effort focused on informative aggregation between views, introducing 
cross-view confidence~\cite{mvrotationnet},
group convolutions to learn rotation-equivariant representations~\cite{mvequivariant},
graph convolutions to learn view aggregation~\cite{mvviewgcn}.
Alternatively, MVTN~\cite{mvtn} predicted the viewpoints from a differentiable 3D renderer.
In the video domain, synthetic views (\eg 3D motion or optical flow) are created for single-stream videos as a way to obtain better representation learned in a self-supervised fashion~\cite{li2018unsupervised,mvvideo}.  
In this work, we leverage a simple multi-view pipeline for video understanding, trained in a fully supervised fashion, that incorporates multiple replay streams from soccer broadcast videos.

\section{SoccerNet-MVFouls dataset}
\label{sec:dataset}

In this section, we introduce our novel multi-view foul classification soccer dataset, called \textit{SoccerNet-MVFouls}.

% \subsection{Overview}
Table~\ref{tab:MOT_Comparison} presents an overview of our dataset and compares it with other datasets that propose action recognition using either single or multiple views.
Our dataset is the only one for multi-view video action recognition in sports, and the first dataset to focus specifically on referee's decisions. 

% \textit{SoccerNet-MVFouls} gathers $8,923$ videos on which we annotated $10$ different properties describing the characteristics of the foul from the point of view of a referee (\eg the severity of the foul, the type of foul, \etc). All annotated properties are listed in Section~\ref{dataset collection} and a detailed description of all properties can be found in supplementary material. % \ref{supp_dataset}).
% To ensure high-quality annotations, all these properties were manually annotated by a professional soccer referee with $6$ years of experience and more than $300$ official games. The referee could watch the videos from all available views at any speed to accurately characterize the foul.
% The dataset is composed of $3901$ actions extracted from $500$ soccer games from six main European leagues, covering three seasons from 2014 to 2017, extracted from the SoccerNet dataset~\cite{Giancola2018SoccerNet,Deliege2021SoccerNetv2}. 
% All videos are available at $25$fps and two resolutions: $720$p or $224$p.
% On average, our dataset contains $2.23$ views per action and a total of $89.230$ annotations.

\textit{SoccerNet-MVFouls} gathers $3{,}901$ actions extracted from $500$ soccer games from six main European leagues, covering three seasons from 2014 to 2017, extracted from the SoccerNet dataset~\cite{Giancola2018SoccerNet,Deliege2021SoccerNetv2}. 
Each action is composed of at least two videos depicting the live action and at least one replay. 
The actions are annotated with $10$ different properties describing the characteristics of the foul from a referee's perspective (\eg the severity of the foul, the type of foul, \etc). 
To ensure high-quality annotations, all these properties were manually annotated by a professional soccer referee with $6$ years of experience and more than $300$ official games. The referee watched the videos from all available views at any speed to accurately characterize the foul.
% All annotated properties are listed in Section~\ref{dataset collection} and a detailed description can be found in supplementary material.

% In the following, we present how we acquired and annotated our multi-view action dataset (see Section~\ref{dataset collection}), followed by a statistical analysis (see Section~\ref{dataset statistics}).
 
\subsection{Dataset collection} \label{dataset collection}

The dataset was collected in three steps: (i)~we extracted the relevant action clips from soccer broadcast videos, (ii)~we temporally aligned the clips related to the same action, and (iii)~we annotated several foul properties.

\begin{figure}[t]
  \centering
  \includegraphics[width=1\linewidth]{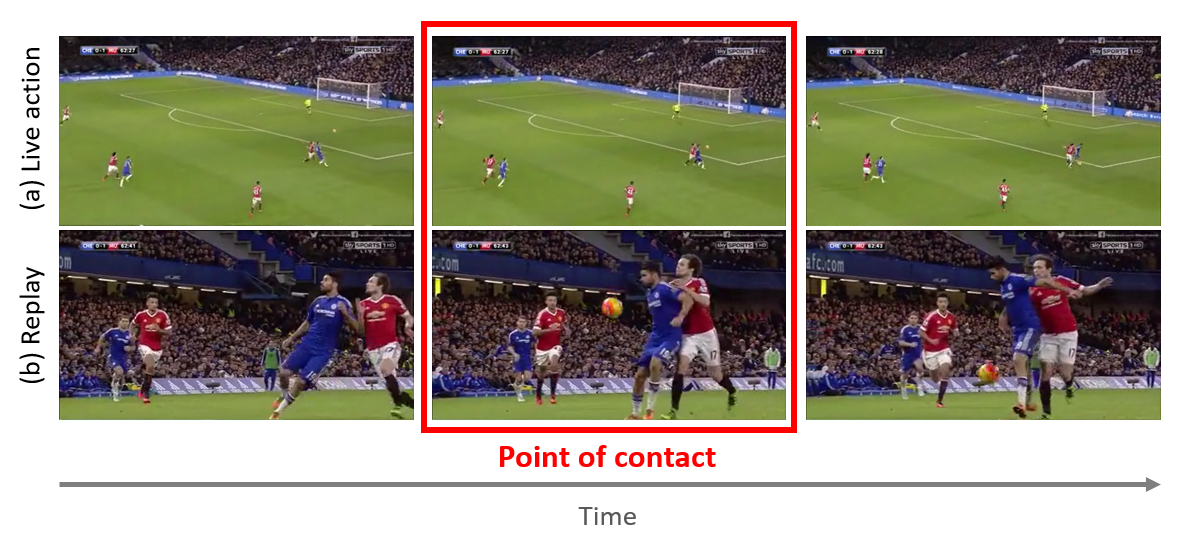}
        \caption{
        \textbf{Example of a multi-view sequence from our dataset.}
        Each foul has at least (a) one live-action clip (usually taken from the main camera) and (b) one synchronized replay clip (usually a close-up view). We annotate the exact frame where the point of contact happens (red box). The ground-truth properties for this example are: ``Offence'', ``Challenge'', ``No card'', ``With contact'', ``Upper body'', ``Use of shoulder'', ``Ball is not played'', ``Tried to play the ball'', ``No handball'', and ``No handball offence''. 
    }
    \label{fig:datset example}
\end{figure}

\mysection{Clip extraction.}
As a starting point, we used the SoccerNet-v2 dataset~\cite{Deliege2021SoccerNetv2}, which contains timestamp annotations of fouls for $500$ full broadcast games.
Furthermore, the SoccerNet dataset also provides annotations of the replays of some of the fouls, allowing us to retrieve, for the same action, different viewpoints. 
Since our goal is to design a multi-view video assistant referee system, we only keep actions for which we have access to at least two different points of view. 
In most cases, the extracted clips should cover sufficient information to determine all the foul properties.
Also, to prevent bias towards the on-field referee's decision, the $5$ second clips should not contain the decision of the referee (\eg if the player is given a yellow card). 
Therefore, we extracted $5$ second clips per action, starting $3$ seconds before and ending $2$ seconds after the timestamp annotation. 
In the following, ``live action clip'' will refer to the clips taken from the main camera, while ``replay clips'' will denote all the replay clips typically taken from closer shots. Figure~\ref{fig:datset example} shows an example of such extracted clip.

\mysection{Clips alignment.}
We build a Multi-View Foul Annotator tool with a similar interface to a VAR room to ensure the quality of the annotations performed by our referee. 
% To ensure the quality of the annotations performed by our referee, we build a Multi-View Foul Annotator with a similar interface to a VAR room. 
At first, the referee is presented with all available clips of an action simultaneously on a grid layout. 
Our annotator tool enables users to modify the annotated point of contact (see Figure~\ref{fig:datset example}) for each clip individually and adjust the speed and offset of the clips to align them temporally, taking into account the fact that replays are frequently broadcasted at a slower speed.
The referee may browse simultaneously the synchronized videos either at regular speed or frame by frame to accurately understand and describe the properties of the action.
More information and an example of annotation using our annotator may be found in the supplementary material.

\mysection{Property annotations.}
The SoccerNet-v2 dataset provides annotations for fouls and yellow/red cards given by the actual game referee.
However, the on-field referee has only his own point of view to characterize the foul.
Judging foul play incidents from the referee's position at playing time leads to an average error rate of $14\%$~\cite{Mallo2012Effect}.
Our referee annotator has no time pressure and access to multiple perspectives, which results in more accurate decisions compared to the on-field referee who has to take a quick decision and only has a single view.
% However, as the on-field referee has only his own point of view of the foul on the field, there is an average error rate of $14\%$ in judging foul play incidents depending on the referee's positioning and the playing time~\cite{Maglo2022Efficient}.
To ensure a high-quality dataset and avoid any bias, our professional soccer referee manually annotated all properties without seeing the on-field referee's decision. % to avoid any bias toward their decision.

We defined several properties for each action that are necessary for the referee to take the final decision. 
These properties include (i) if the clip contains a foul (\ie an action which breaks/violates the \textit{Laws of the Game}~\cite{IFAB2022Laws}), (ii) the class of the foul, (iii) the severity of the foul, (iv) if the player plays the ball, (v) if the player tries to play the ball, (vi) if any player touches the ball with his hand or arm, deliberately or not, (vii) and whether it is an offence according to the \textit{Laws of the Game}~\cite{IFAB2022Laws}), (viii) if there is contact between two players, (ix) the action foul relates to the upper or underbody, and, finally, (x) we further discriminate for the upper body between arms and shoulders.
We have special labels corresponding to grey areas for the property (i), we use the label ``Between'' when both ``Foul'' and ``No foul'' decisions are equally valid and there is no obvious decision.
For property (iii), we use the labels ``Borderline No card/Yellow card'' and ``Borderline Yellow card/Red card'' to indicate a grey area when either ``No card'' or ``Yellow card', (resp. ``Yellow card'' or ``Red card''), would be the correct decision.
% More details can be found in the supplementary material. 
% A detailed description of all the annotated properties can be found in the supplementary material. 

%The final step consisted in time aligning the clips of the same foul and to adjust the speed of the replays to match real-time speed.
%Unfortunately, not all replays were slow down by the same factor and we had to go over all the clips and check them individually.  

%An example of a multi-view clip with its corresponding groundtruth can be seen in Figure \ref{fig:datset example} (more examples in the annexe \ref{dataset example annexe}).

\subsection{Dataset statistics} \label{dataset statistics}

\mysection{Number of views.}
% For around $75\%$ of the fouls, we have two different viewpoints where one view corresponds to the live action and one view is the corresponding replay. 
% Furthermore, we have $20\%$ of the fouls where we have three different viewpoints per foul, corresponding to one live action and two replays.
% Finally, we have approximately $5\%$ of the fouls for which we have $4$ different viewpoints, corresponding to one live action and three replays. 
% In average, we have $2.29$ clips per foul. 
% 
On average, we have $2.29$ clips per foul action, around $75\%$ of them have two viewpoints (live and replay), $20\%$ have a second replay, and around $5\%$ have a third replay video. No foul has more than four views.

% Action classification: Performance per class

\begin{table}[t]
    \centering
    \resizebox{\linewidth}{!}{% <------ Don't forget this %
    \begin{tabular}{lc|lc|lc}
    \toprule
        \multicolumn{2}{c|}{\bf Fouls} &  \multicolumn{2}{c|}{\bf Offence} &  \multicolumn{2}{c}{\bf Offence Severity} \\ \midrule
       \bf Class & \bf Prob. & \bf Class & \bf Prob. & \bf  Class & \bf Prob.\\ \midrule
        St. tackling & 43.6   &Offence  &85.8    & No card     &55.3\\ 
        Tackling     & 15.6   &No offence  &10.7 & Yellow card &26.6\\
        Challenge    & 13.0   &Between  &3.4     & NC/YC       &15.2\\ 
        Holding      & 12.5   &      &           & YC/RD       &1.7 \\
        Elbowing     & 5.9    &      &           & Red card    &1.1 \\
        High leg     & 3.5  &  &  &  & \\
        Pushing      & 2.9  &  &  &  & \\
        Dive         & 0.9  &  &  &  & \\
    \bottomrule
    \end{tabular}
    }
    \caption{
    \textbf{Distribution of classes in our SoccerNet-MVFouls dataset.}
    The distribution of the classes for the ``Offence'', ``Severity'' and ``Type of foul'' properties is highly imbalanced. The distribution for the other properties is shown in supplementary. ``St. tackling'' stands for standing tackling.
    }
    \label{tab:distribution}
\end{table}

 %  \begin{figure}[t]
 %    \centering
 %    \includegraphics[width=1\linewidth]{figures/Dataset/class_distribution.png}
    %           \caption{\textbf{Distribution of classes in our SoccerNet-MVFouls dataset} The distribution of the classes for the ``Offence'', ``Severity'' and ``Type of foul'' properties is highly imbalanced. The distribution for the other properties are shown in supplementary material.\AH{including the in between in the dataset description and then removing them in the benchmark is a bit weird }}
   %        \label{distribution}
%   \end{figure}

\mysection{Properties distribution.}
Table~\ref{tab:distribution} shows the distribution of the properties ``Offence'', ``Severity'' and ``Type of foul'.
We can see that in all three cases, the distribution is highly unbalanced towards ``No card'', ``Offence'' and ``Standing tackling'', respectively.
% The most represented foul corresponds to an ``Offence'', where the type of foul is a ``Standing tackling'' and the offence resulted in ``No card'' given to the player.
This analysis follows our intuition of soccer, where yellow and red cards are usually rarer than simple free-kicks given after a foul.

\begin{table}[t]
    \centering
    \resizebox{\linewidth}{!}{% <------ Don't forget this %
    \begin{tabular}{l|c|ccc}
    \toprule
    && \multicolumn{3}{c}{\textbf{Severity Distribution}} \\
 \textbf{Foul Class} & \textbf{Succ. Rate} &\textbf{No Card} &\fcolorbox{black}{yellow}{\rule{0pt}{4pt}\rule{4pt}{0pt}}~~ \textbf{Card} &\fcolorbox{black}{red}{\rule{0pt}{4pt}\rule{4pt}{0pt}}~~ \textbf{Card} \\\midrule
Standing Tackling  &\textbf{0.94} &$0.79$ &$0.18$ &$0.02$ \\
Tackling  &$0.87$ &$0.37$ &\textbf{0.58} &\textbf{0.04} \\
High Leg  &$0.87$ &$0.31$ &\textbf{0.63} &\textbf{0.06} \\
Holding  &$0.90$ &$0.60$ &$0.40$ &$0.00$ \\
Pushing  &$0.84$ &\textbf{0.99} &$0.01$ &$0.00$ \\
Elbowing &\textbf{0.93} &$0.43$ &\textbf{0.53} &$0.03$ \\
Challenge &$0.75$  &\textbf{0.94} &$0.05$ &$0.01$ \\
Dive  &/ &$0.00$ &$1.00$ &$0.00$ \\ \bottomrule

    \end{tabular}
    }
    \caption{\textbf{Referees success rate and severity per foul class.} 
    Referees are successful in most classes but struggle with ``Challenge''. Some classes are less likely to return a card, \eg ``Tackling'' or ``High leg''. 
    The success rate for ``Dive'' is unknown, as we cannot know if a referee whistle for the foul or the dive.
    % It is not possible to provide the ``Dive'' statistic as we do not know if the referee whistled for the foul or the dive.
    % For most classes, the referees have a very high success rate. However, they have difficulties in correctly classifying ``Challenge'' as foul or no foul. Let us note that it is not possible to provide the ``Dive'' statistic as we do not know if the referee whistled for the foul or the dive. 
    % %Furthermore, the foul class is strongly correlated with the severity.
    % For some classes, the referees rarely or never give a yellow card whereas for other foul classes such as ``Tackling'' or ``High leg',  there is a high probability for a player to receive a yellow card.
    \label{tab:referee_success_rate}}
\end{table}

\mysection{Success rate of the referees.}
As we only have extracted fouls for which the on-field referee has given a foul in the game, we can analyze the success rate of the referees by analyzing the property ``No offence''.
%We go one step further and we analyze the percentage of correct decisions for each action class.
From the $3901$ fouls given by the referees in the games, our referee annotated $368$ fouls as ``No offence'', leading to an error rate of $10.7\%$. 
``Standing tackling'' and ``Elbowing'' are the most well classified with $94\%$ success rate, as shown in Table~\ref{tab:referee_success_rate}.
For the remaining action classes, the referees have a similar success rate of approximately $87\%$, except for the foul class ``Challenge'', where the referees have an error rate of $25\%$. 
Our analysis is aligned with the finding of Mallo~\etal~\cite{Mallo2012Effect}.
% Our analysis gives similar results to the work of Mallo~\etal~\cite{mallo2012effect}.

\mysection{Severity for different foul classes.}
The distribution of the severity among different foul classes can provide insight into how often certain types of fouls result in a card. %being shown to the offending player. 
The results are presented in Table~\ref{tab:referee_success_rate}.
``Tackling'', ``High leg'', and ``Elbowing'' are three types of fouls that very often result in a yellow card, as they represent fouls that are dangerous for opponents.
Contrarily, some classes like ``Pushing'' or ``Challenge'' are very unlikely to get a yellow or red card.

%% Action classification: Performance per class

\section{Methodology}
\label{sec:method}

% WITH TASK IN DATASET
% We propose a Video Assistant Referee System (VARS) that solves the proposed tasks by automatically classifying the multi-view videos into their respective classes of fouls and offences.

% WITH TASK IN METHODOLOGY

\begin{figure}[t]
    \centering
    \includegraphics[width=\linewidth]{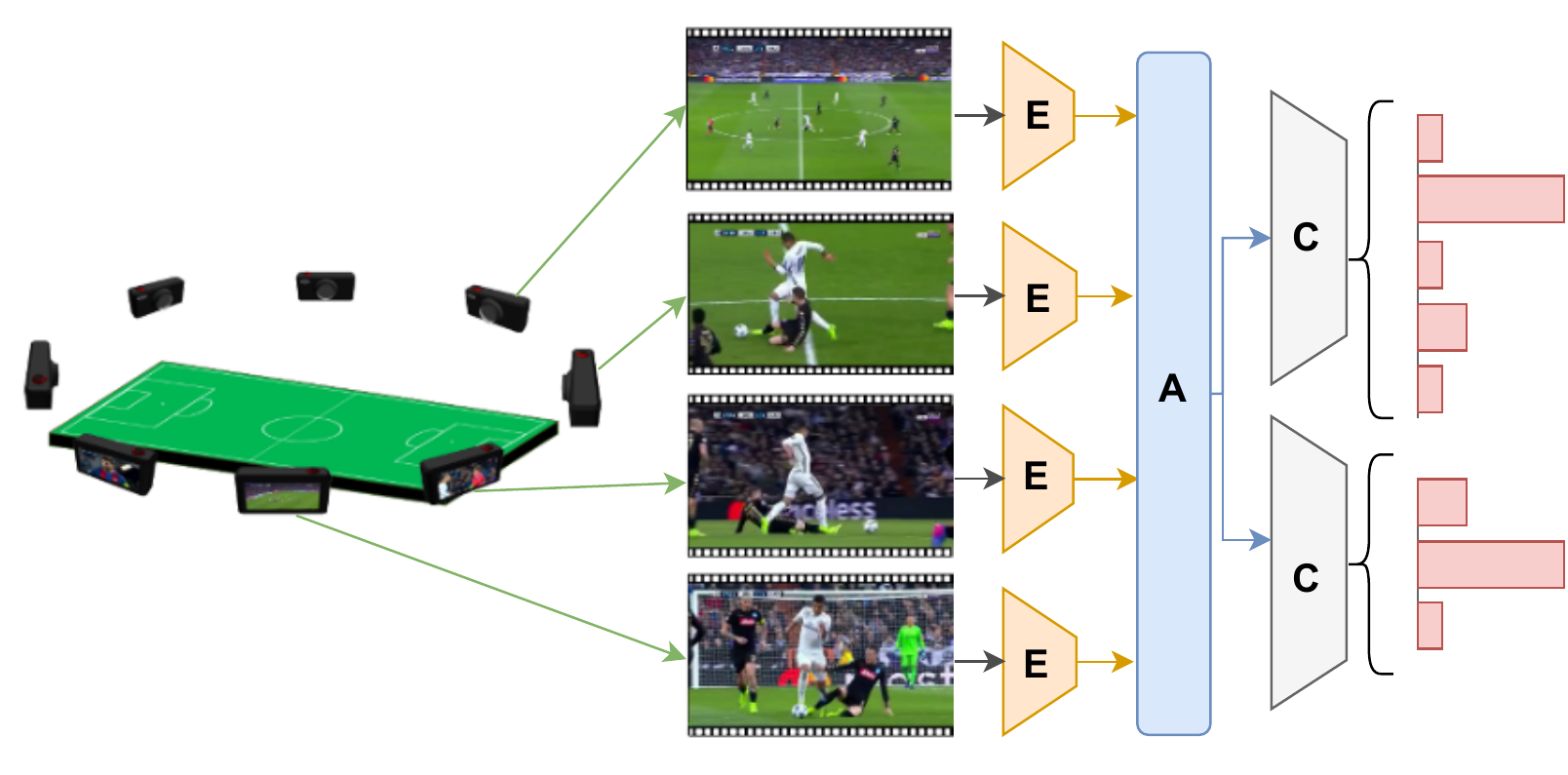}
    \caption{
    \textbf{VARS: Video Assistant Referee System.}
    From multi-view video clips input, our system
    \textcolor{Dandelion}{encodes per-view video features ($\mathbf{E}$)},
    \textcolor{blue}{aggregates the view features ($\mathbf{A}$)}, and
    \textcolor{black}{classifies different properties of the foul action ($\mathbf{C}$)}.
    % Our system takes as input video clips taken from different point of views of the soccer field and creates one feature per clip by passing them through a \textcolor{Dandelion}{video encoder $\mathbf{E}$}. The clip feature representation are then passed to a \textcolor{blue}{multi-view aggregation operation $\mathbf{A}$} that outputs a single feature representation passed through a \textcolor{Gray}{classification head $\mathbf{C}$} to produce the \textcolor{red}{per-class predictions}.
    }
    \label{fig:pipeline}
\end{figure}

Our VAR system is a multi-view video architecture, that automatically identifies different properties for an action.
We illustrate our proposed VARS in Figure~\ref{fig:pipeline}. 
% In the following, we define the different tasks of the VARS~\ref{label:classification task} and the architecture of our system~\ref{label:VARS}.

\subsection{Classification tasks}\label{label:classification task}

We formally define two tasks for our dataset. 

\mysection{Task 1: Fine-grained foul classification.}
Given multiple clips of the same foul instance, the objective is to classify the foul into one of $8$ fine-grained foul classes: ``Standing tackling'', ``Tackling'', ``High leg'', ``Pushing'', ``Holding'', ``Elbowing', ``Challenge'', ``Dive/Simulation''.

% foot fouls ('Standing tackling', 'Tackling', 'High leg'), illegal use of arms ('Pushing', 'Holding', 'Elbowing'), use of shoulder ('Challenge'), and unsporting behavior ('Dive/Simulation').

\mysection{Task 2: Offence severity classification.}
Given multiple clips of the same foul instance, the objective is to classify whether the foul constitutes an offence, as well as the severity of the foul. 
We have defined four classes: ``No offence'', ``Offence + No card'', ``Offence + Yellow card'', and ``Offence + Red card''.
We put aside clips labeled ``Between'' as well as the clips annotated as ``Borderline''.
Therefore, for this particular task, we use a subset of our SoccerNet-MVFoul dataset.
% For the following experiments, we, therefore, use a subset of our SoccerNet-MVFoul dataset

\subsection{Video Assistant Referee System (VARS)} \label{label:VARS}

We propose a novel Video Assistant Referee System (VARS) for the task of video recognition from multiple camera views. 
The pipeline of the VARS is presented in Figure~\ref{fig:pipeline}.
Our VARS takes multiple video clips denoted by $\mathbf{v} = \{v_i\}_{1}^n$ as input, showing the same action from $n$ different views. 
% Our VARS takes multiple video clips as input a vector $ \mathbf{v}$ containing n clips from different views $v_{i=1}^n$ showing the same action. 
A video $v_i$ is fed into a video encoder $\mathbf{E}$ with parameters $\theta_E$ to extract a vector $f_i$ containing the spatio-temporal features for that specific view:  
\begin{equation}
    % \mathbf{F} = \mathbf{E_{\theta_E}}(\mathbf{v}) \, .
    f_i = \mathbf{E_{\theta_E}}(v_i) \ .
\end{equation}
We aggregate the feature vectors through a function $\mathbf{A}$ that outputs a single multi-view representation $\mathbf{R}$ following:

\begin{equation}
    \mathbf{R} = \mathbf{A}(\{{f_i\}_{i=1}^n}) \ ,
\end{equation}
with $\mathbf{A}$ being a max or mean aggregation function.
For the single-task classifier, we input the pooled features through a classification head $\mathbf{C}$ with parameters $\theta_C$.
% From the output probability vector of the classification head, VARS selects the maximum value as its final prediction as given by
VARS predicts the final class from the maximum probability score of the classification head, as given by:

\begin{equation}
    \mathbf{VARS} = \arg\max \mathbf{C_{\theta_C}}(\mathbf{R}) \ .
\end{equation}
We train our model to minimize the following loss:

\begin{equation}
    \mathcal{L} = 
    \mathbf{L}(\mathbf{C_{\theta_C}}\
    (\mathbf{A}
    ( \{\mathbf{E_{\theta_E}}(v_i) \}_{i=1}^n )), y) \, ,
\end{equation}
with $\mathbf{L}$ being the cross entropy loss function, and $y$ the ground truth associated to $\{v_i\}_{i=1}^n$.
For the offence severity classification, VARS has to understand the game of soccer in order to correctly classify fouls into ``No card'', ``Yellow card'', and ``Red card''.
In fact, bringing contextual information about the type of foul inside the network is essential to determine the severity of the offence. % which can influence the decision of whether or not to give a card.
As the foul and offence classifiers share common features, we train a model to perform both tasks simultaneously.
Our multi-task VARS learns to leverage these shared features to improve its predictions for both tasks.
For the multi-task classifier, 
we define two heads,  $\mathbf{C^{foul}}$ and $\mathbf{C^{off}}$, respectively for the tasks of fine-grained foul classification and offence severity classification.
% we have a classification head per task $\mathbf{C_t}$ with parameters $\theta_{C_t}$. %with the same architecture as previously with the difference that the output dimension corresponds to the number of classes for task t.
From the probability vector of each task, the VARS will take the maximum as the final prediction:
\begin{equation}
    \mathbf{VARS^{t}}= \arg \max \mathbf{C^t_{\theta_{C^t}}}(\mathbf{R}) \quad \textrm{$\forall \mathbf{t} \in \{\text{foul}, \text{off}\}$} \, .
\end{equation}
We train our model by minimizing both tasks loss with:
%For our training set of size s, we want to minimize:
\begin{equation}
    % \underset{\theta_E, \theta_C} {\min} \sum_{i=0}^s 
    \alpha_\text{foul} \mathcal{L}^\text{foul} + \alpha_\text{off} \mathcal{L}^\text{off} \, .
    % \underset{\theta_E, \theta_C} {\min} \sum_{i=0}^s \sum_{j=0}^t \alpha_j \mathbf{L_j}\ ,
\end{equation}
% where:
% \begin{equation}
%     \mathbf{L_j} =   \mathbf{L}(\mathbf{C_{j,\theta_{C_j}}}\
%     (\mathbf{A}
%     ( \mathbf{E_{\theta_E}}(\textbf{v})))) \, .
% \end{equation}

%with $\mathbf{L}$ the cross entropy loss function and $\mathbf{L_j}$ the cross entropy loss for the task j.
By choosing different values for $\alpha$, we can assign more or less importance to tasks. 
This scaling is necessary when the losses have significantly different magnitudes. 
In the case of our two tasks, the losses have a similar order of magnitude, so we typically select $\alpha_{foul}=\alpha_{off}=1$.

\mysection{Video encoder $\mathbf{E}$.}
We considered different encoders to extract features from the video clips: 
\textbf{(i)} \underline{ResNet}~\cite{he2016deep} may be used on videos by running the network on each frame independently and then using a max or mean pooling operation on the features across the frames to obtain a single feature vector that represents the entire video. While this approach works well for extracting spatial features, it does not capture temporal dynamics. % of the video.
\textbf{(ii)} \underline{R(2+1)D}~\cite{tran2018closer} extends the 2D CNN architecture with an additional temporal convolutional layer that operates on a sequence of frames to capture the temporal dynamics of the video. The advantage compared to ResNet is that it both captures spatial and temporal features directly.
\textbf{(iii)} \underline{MViT}~\cite{li2022mvitv2, fan2021multiscale} integrates a multiscale feature representation with a transformer-based architecture to capture both spatial and temporal information from video clips.
The feature encoders are typically pre-trained on ImageNet~\cite{deng2009imagenet} (ResNet) or Kinetics~\cite{Kay2017TheKinetics-arxiv} (R2+1D and MViT).

\mysection{Multi-view aggregator $\mathbf{A}$.}
To combine the extracted features from multiple views, we introduce two different pooling strategies~\cite{hamdi2022mvtn}, in particular: 
\textbf{(i)} \underline{Mean pooling} takes the average value for each feature, and 
\textbf{(ii)} \underline{Max pooling} which takes the maximum value per feature. 

\mysection{Classification heads $\mathbf{C}$.}
%Our classification heads are responsible for transforming the aggregated features into a probability distribution over the classes. 
Our classification heads consist of two dense layers with softmax activation. 
The output is a probability vector with dimensions that match the number of classes in the classification problems.

%Statistics about the frequency of different types of fouls can provide valuable information for referees to better prepare for games. 
%By analyzing this data, referees can identify which types of fouls are most likely to occur during a game and he can focus his attention on those areas.
%However, manually annotating fouls during games is a costly and time-consuming process.

\section{Experiments} \label{sec:benchmarks}

\subsection{Experimental setup}

\mysection{Training details.}
For both classification tasks, we leverage clips of $16$ frames, spanning temporally for $1$ second, with a spatial dimension of $224\times398$ pixels.
Specifically, the clips contain $8$ frames before the foul and $8$ frames after the foul.
The encoders $\mathbf{E}$ are pre-trained as detailed in the methodology, and the classifier $\mathbf{C}$ is trained from scratch, while both are trained in an end-to-end fashion. 
We use a cross-entropy loss, optimized with Adam with an exponential decreasing learning rate starting at $10^{-4}$ and a batch size of $8$.
The model starts overfitting after $10$ epochs, and it takes around $9$ hours to train on a single Nvidia V100 GPU.

% The trainings take around $9$ hours. 

\mysection{Evaluation metrics.}
We report the classification accuracy, which is defined as the ratio of actions correctly classified with respect to the total number of actions.
We also provide the top-2 accuracy (where a sample is considered well classified if the class appears in the top two highest confidence predictions) to get more insight into the model's performance.
As our dataset is unbalanced, we also provide the balanced accuracy, which is defined as follows:

\begin{equation}
    \textrm{\mbox{Balanced Accuracy (BA)}} = \frac{1}{N}\sum_{i=1}^{N} \frac{TP_i}{P_i} \, ,
\end{equation}
with \textit{N} the number of classes, $TP$ (True Positives) is the number of times where the model correctly predicted the class $i$ and $P_{i}$ (Positives) is the number of ground-truth samples for that class in the dataset.

\subsection{Main Results}

\mysection{Task 1: Fine-grained foul classification.}
Our results may be found in Table~\ref{tab:AblationStudyActionClassification}.
By extracting spatio-temporal features with MViT, we achieve significant improvements in performance compared to ResNet and R(2+1)D. 
This indicates that using a more advanced feature encoder can significantly enhance the model's ability to identify and classify the type of foul.
The influence of the pooling method on the performance is however not significant, although max pooling shows slightly better results.
%Regarding the feature encoder, we found that R(2+1)D and ResNet show similar performance, despite R(2+1)D taking temporal features into account while ResNet only considers spatial features. 
%However, by using MViT, we achieve significant improvements in performance. 
%max-pooling tends to provide slightly better results for the accuracy than mean-pooling. 
%For the balanced accuracy, mean-pooling has a slightly better performance compared to max-pooling. 
In general, max pooling might be better when not all views are equally informative.
Taking the max values helps identify the most important features for the most informative views while ignoring less useful information. 
In contrast, mean-pooling takes into account the information from all views, including those with a poor perspective. 
%Specifically, we observed an improvement of 0.9 for mean-pooling and 0.15 for max-pooling when compared to R(2+1)D and ResNet.
%In contrast, the choice of pooling method has only a minor impact on the model's performance.
Overall, the best performance is obtained by using MViT as video encoder and max pooling.%, with an accuracy of 0.47 and a balanced accuracy of 0.43.

\begin{table}[t]
    \centering
    \resizebox{0.92\linewidth}{!}{% <------ Don't forget this %
    \begin{tabular}{ll|ccc}
    \toprule
\textbf{Feature Extractor} & \textbf{Pooling}  & \textbf{Acc. @1} & \textbf{Acc. @2} &\textbf{BA} \\\midrule
ResNet \cite{he2016deep} &Mean &$0{.}31$ &$0{.}56$ &$0{.}28$ \\
ResNet \cite{he2016deep} &Max &$0{.}32$ &$0{.}60$ &$0{.}28$  \\
R(2+1)D \cite{tran2018closer} &Mean  &$0{.}31$ &$0{.}55$ &$0{.}34$ \\
R(2+1)D \cite{tran2018closer} &Max  &0.32   &0.56 &$0{.}33$ \\
MViT \cite{li2022mvitv2, fan2021multiscale} &Mean   &$0{.}40$ &$0{.}65$ &\textbf{0.45}   \\
MViT \cite{li2022mvitv2, fan2021multiscale}   & Max  &\textbf{0{.}47} &\textbf{0{.}69} &$0{.}43$\\
\bottomrule
    \end{tabular}
    }
    \caption{ \textbf{Main results for the multi-view video foul classification.}  We compare three feature encoders and two pooling methods. The best performance is obtained with MViT and a max pooling between the views. BA indicates the balanced accuracy after normalizing by the frequency of that class.}
    \label{tab:AblationStudyActionClassification}
\end{table}

\begin{table}[t]
    \centering
    \resizebox{0.92\linewidth}{!}{% <------ Don't forget this %
    \begin{tabular}{l|cc|cc} 
    \toprule
\textbf{Feature Extractor} & \textbf{Pooling} &\textbf{Task}  & \textbf{Acc.} & \textbf{BA} \\\midrule
ResNet \cite{he2016deep} &Mean &Single &$0{.}25$  &$0{.}26$  \\
ResNet \cite{he2016deep} &Max &Single &$0{.}22$ &$0{.}25$ \\
R(2+1)D \cite{tran2018closer} &Mean &Single &$0{.}28$ &$0{.}30$ \\
R(2+1)D \cite{tran2018closer} &Max &Single &$0{.}27$  &$0{.}29$ \\
MViT \cite{li2022mvitv2, fan2021multiscale} &Mean & Single  &$0{.}32$ &$0{.}23$     \\
MViT \cite{li2022mvitv2, fan2021multiscale}   & Max & Single &$0{.}29$ &$0{.}27$ \\ \hline
ResNet \cite{he2016deep} &Mean & Multi  &$0{.}34$ &$0{.}25$ \\
ResNet \cite{he2016deep} &Mean & Multi  &$0{.}32$ &$0{.}24$ \\
R(2+1)D \cite{tran2018closer} &Mean &Multi &$0{.}34$ &$0{.}30$ \\
R(2+1)D \cite{tran2018closer} &Max &Multi &$0{.}39$  &$0{.}31$\\
MViT \cite{li2022mvitv2, fan2021multiscale} &Mean & Multi  &$0{.}38$ &$0{.}31$ \\
MViT \cite{li2022mvitv2, fan2021multiscale} &Max & Multi  &\textbf{0{.}43} &\textbf{0{.}34} \\ \bottomrule
    \end{tabular}
    }
    \caption{\textbf{Multi-view video offence and severity classification}. We evaluate our VARS with different feature encoders and pooling methods on a single and multi-task setup. BA stands for the balanced accuracy.}
    \label{tab:AblationStudyOffenceSeverityClassification}
\end{table}

\begin{figure}
\centering
\begin{subfigure}{0.23\textwidth}
    \includegraphics[width=\textwidth]{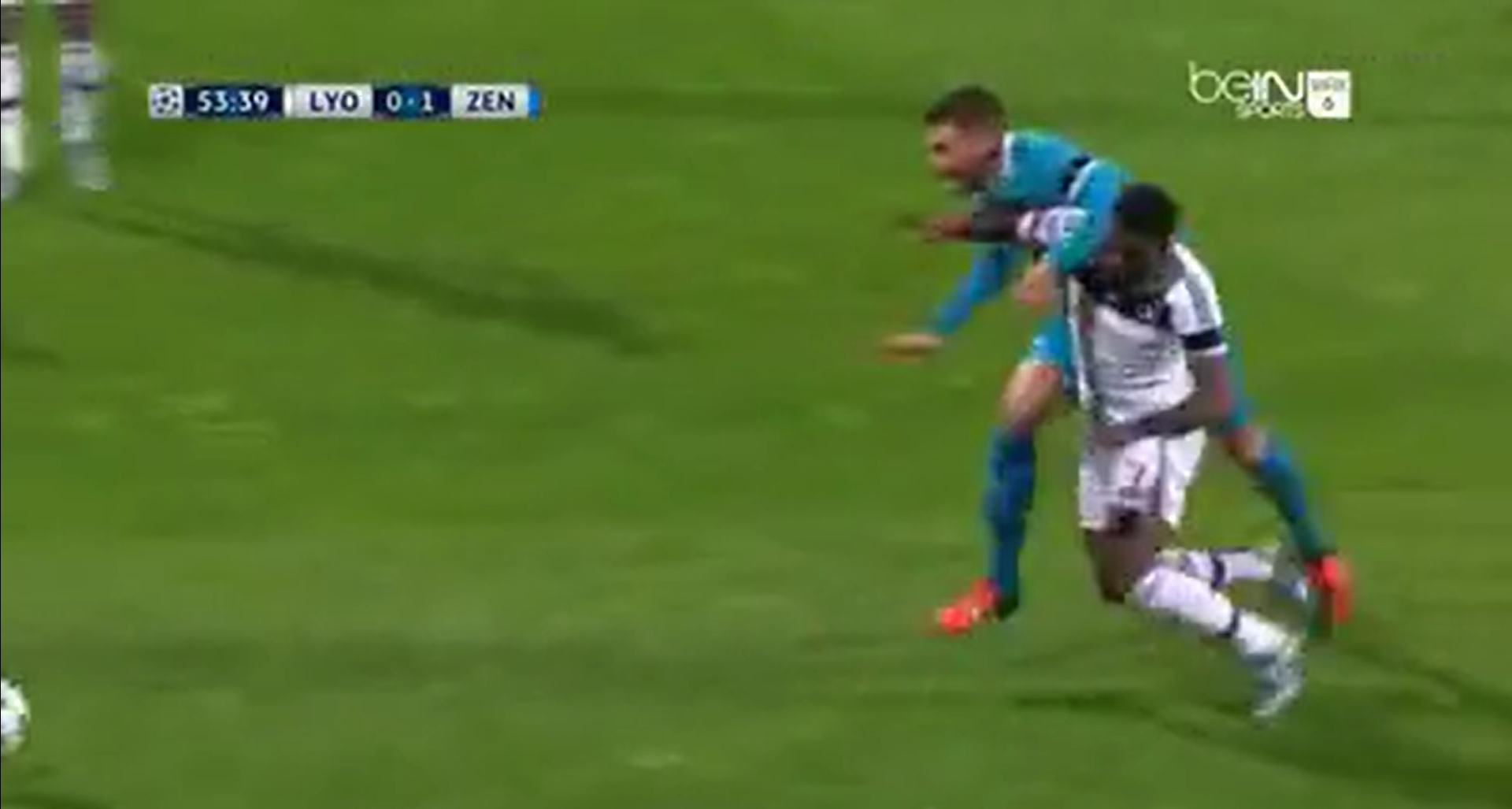}
    \caption{Elbowing - Yellow card}
    \label{fig:illegalUseOfArms}
\end{subfigure}
\hfill
\begin{subfigure}{0.23\textwidth}
  \includegraphics[width=\textwidth]{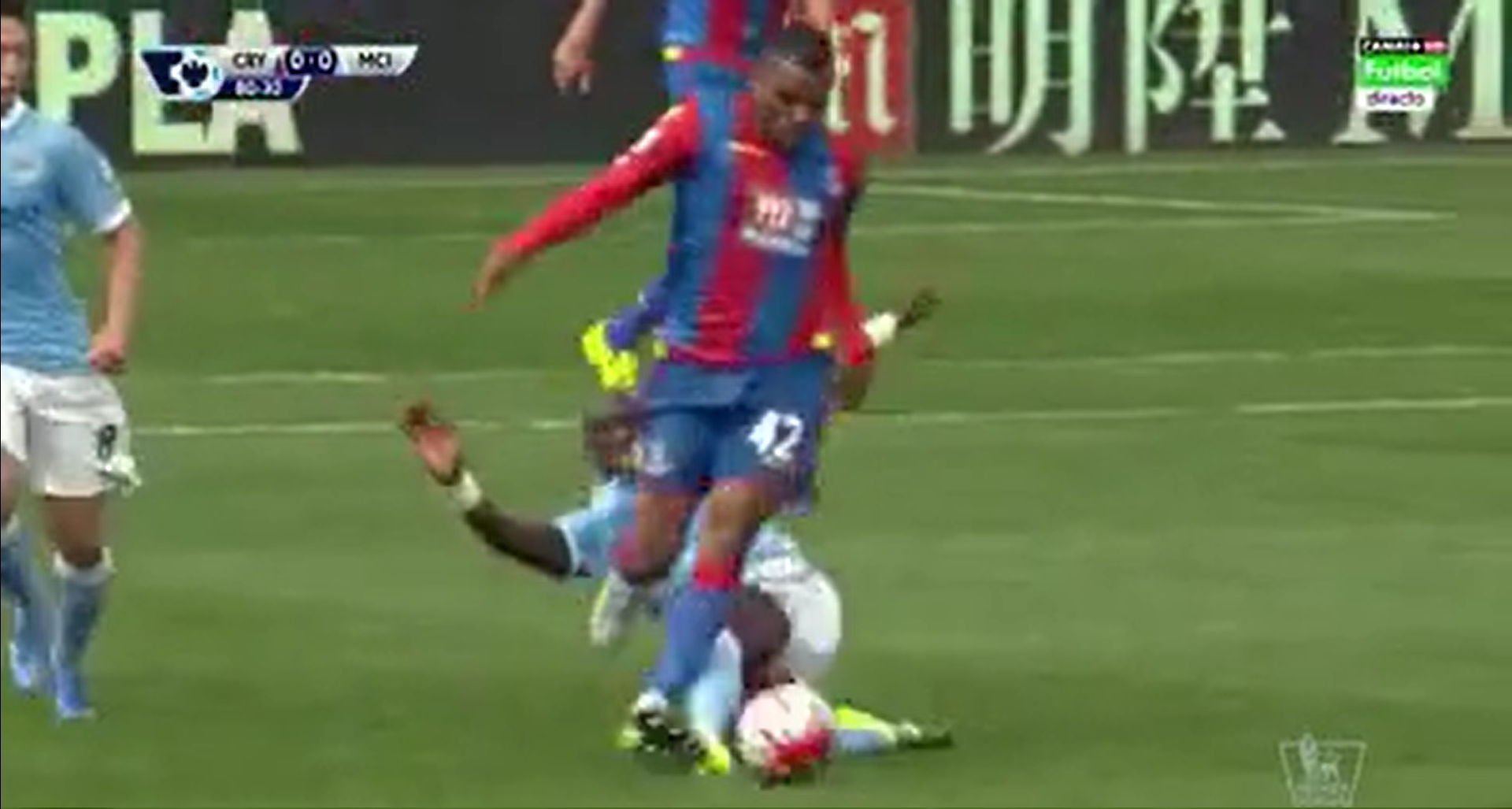}
    \caption{Tackling - Yellow card}
    \label{fig:tackling}
\end{subfigure}
\hfill
\begin{subfigure}{0.23\textwidth}
    \includegraphics[width=\textwidth]{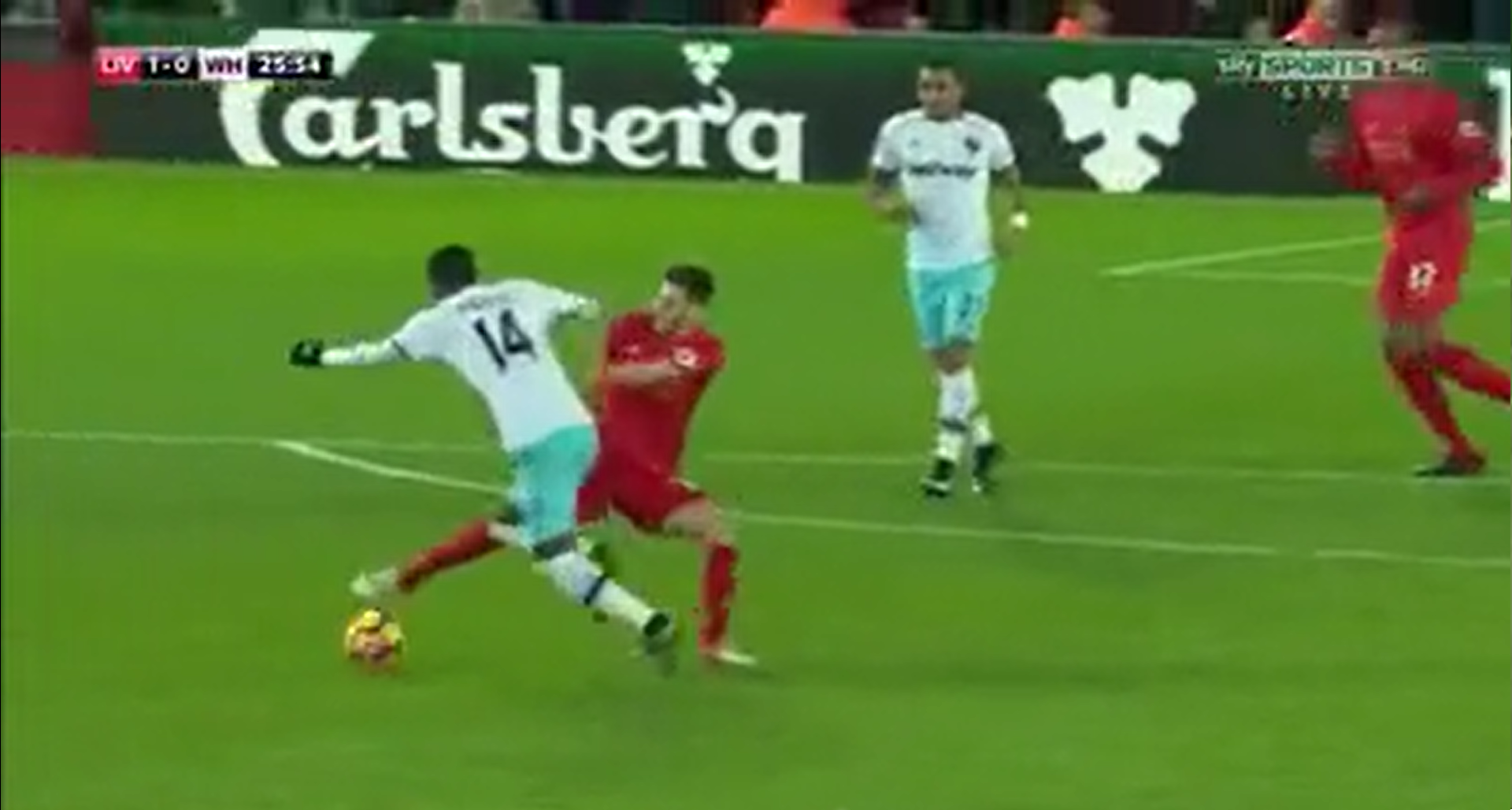}
   \caption{Standing Tackling - No card}
    \label{fig:no card}
\end{subfigure}
\hfill
\begin{subfigure}{0.23\textwidth}
    \includegraphics[width=\textwidth]{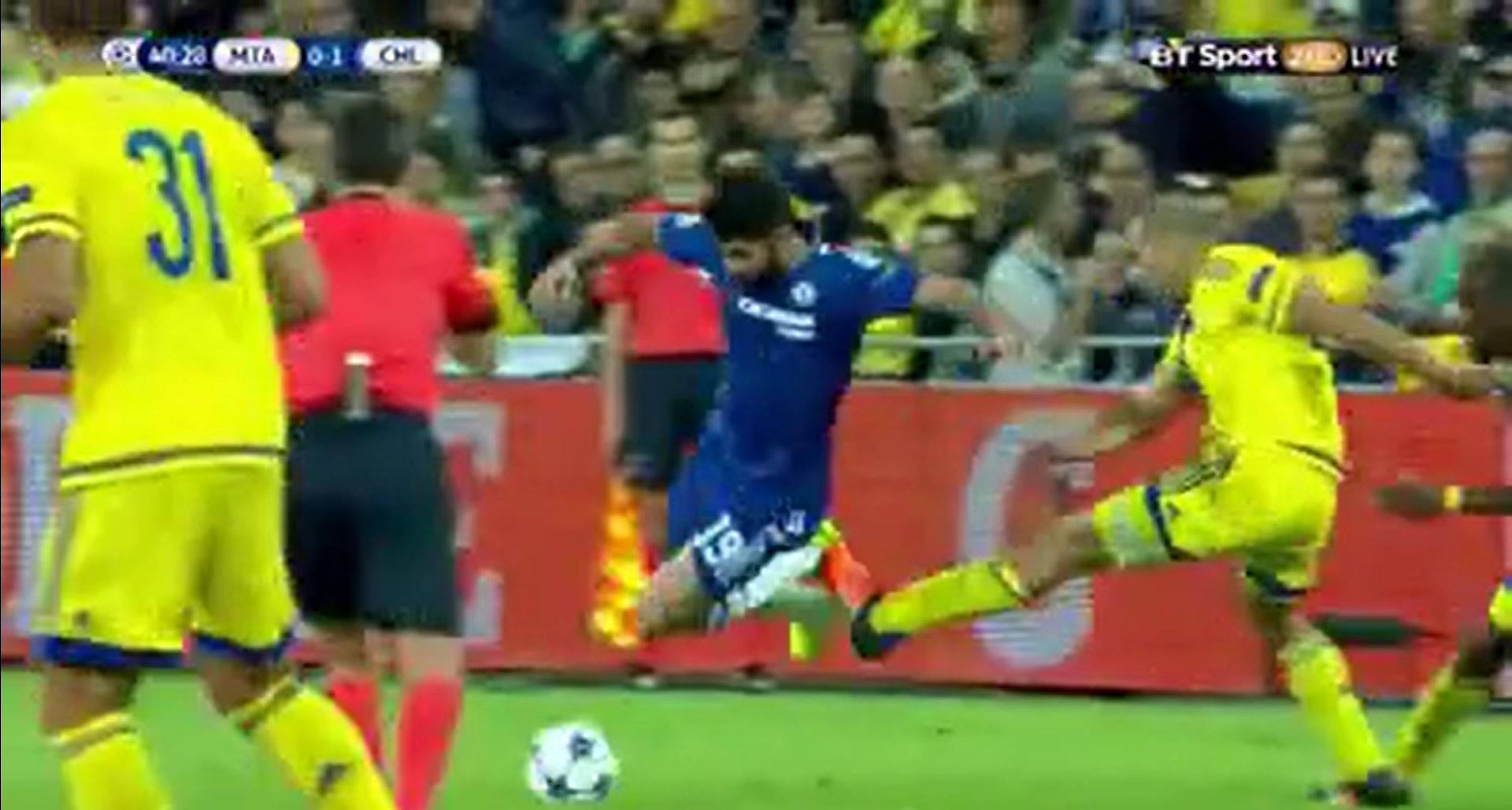}
    \caption{Standing Tackling - Red card}
    \label{fig:red card}
\end{subfigure}
\caption{
\textbf{Example of fouls.}
% \textbf{Comparison between two fouls resulting both in a yellow card.} 
(a) The defender uses his arm as a tool to gain an unfair advantage and ignores the potential danger for his opponent. 
(b) The defender makes a tackle while taking the risk of his opponent being injured. 
(c) The defender tries to play the ball in no dangerous way. 
(d) The defender has no intention to play the ball and only aims to harm his opponent.}
 \label{similar visuall offence severity}
\end{figure}

\mysection{Task 2: Offence severity classification.}
% \label{OffenceClassification}
For the offence severity classification, we study the same feature encoders and pooling techniques. 
The top part of Table~\ref{tab:AblationStudyOffenceSeverityClassification} shows the results obtained by our single-task classifier. 
Regardless of the used feature extractor or pooling technique, the model has more difficulties in classifying the actions. 
%The model is not able to generalize during training and the validation loss stayed the same while the validation accuracy fluctuate around roughly 27\%. 
% 
These difficulties are mainly due to two factors.
First, the dataset exclusively consists of actions that were awarded a free kick by the on-field referee. 
%As a result, the ``No offence'' actions are instances that are visually similar to a foul, and no clear ``No offence'' actions.
As a result, the ``No offence'' actions are visually similar to a foul, and not to clear ``No offence'' actions.
The model often struggles to differentiate these actions from actual fouls, 
which can be further seen in the Supplementary Material.
% which can be seen in the confusion matrix shown in Figure~\ref{confusion_matrix_offence_main}.
Secondly, the visual appearance of an offence with no card, yellow card, or red card can vary greatly. 
In Figures~\ref{fig:illegalUseOfArms} and~\ref{fig:tackling}, we compare two frames of two different foul classes that have a little visual similarity. 
However, in both cases, the defender acted with disregard for the safety of his opponent and therefore resulting in a yellow card.
In contrast, Figures~\ref{fig:no card} and~\ref{fig:red card} depict fouls that are visually more similar than the previous two fouls, yet one resulted in ``No card'' while the other resulted in a ``Red card''.
Minor differences such as the point of contact, the speed of the foul, the distance to the ball, and the intention to play the ball or not, can lead to different classifications.
%This shows that this new task that we introduce is challenging and would require more investigation in the future.
%In the supplementary material, we provide a more thorough analysis of the training performance.

\mysection{Multi-task classifier.} Training a multi-task classifier on related tasks allows the model to utilize the learned information from one task to improve the performance on other tasks. 
In the bottom part of Table~\ref{tab:AblationStudyOffenceSeverityClassification}, we can see that the multi-task classifier outperforms the single-task classifier regardless of the feature encoder or pooling technique for offence severity classification.
Using ResNet to extract spatial features for the type of foul and the offence severity classification does not perform well for either task. The body movements over time and the speed of the players involved in an action are important factors that can greatly impact the outcome of the classification. 
The multi-task classifier combined with MViT as encoder and max pooling shows promising results in classifying actions into their corresponding offence severity class.
Furthermore, the multi-task classifier shows similar results as obtained for the single-task type of foul classification.

\begin{table}[t]
    \centering
    \resizebox{\linewidth}{!}{% <------ Don't forget this %
\begin{tabular}{l|ccccc}
\toprule
 \multirow{2}{*}{{\textbf{Performance}}} & \multicolumn{5}{c}{\textbf{Viewing Setup}} \\
 & L & R1 & L+R1 & R1+R2 & L+R1+R2 \\ \midrule
Acc$_{T1}$     & 0.31 &  0.47  &   0.50  &   0.56 &\bf 0.57 \\
Acc$_{T1}$@2   & 0.54 &  0.68  &   0.70  &   0.69 &\bf 0.72 \\ 
BA$_{T1}$   & 0.29 &  0.38  &   0.36  &\bf0.44 &    0.39 \\ \midrule
Acc$_{T2}$     & 0.38 &  0.39  &\bf0.43  &   0.39 &    0.40 \\
Acc$_{T2}$@2   & 0.67  & 0.70   &  0.72  &   0.73 &\bf 0.75  \\
BA$_{T2}$   & 0.38 &  0.27  &   0.34  &   0.27 &\bf 0.39 \\ \bottomrule
\end{tabular}
    }
    \caption{ 
    \textbf{Single \vs multi-view classification}. 
    We compare the performance for single vs multi-views and the influence of the type of view (Live \textit{L} and replay \textit{R}). We use MViT~\cite{li2022mvitv2, fan2021multiscale} as feature extractor and max pooling. For both tasks, the best performance is mostly obtained with all three views. BA stands for balanced accuray, T1 stands for task 1 (foul classification) and T2 stands for task 2 (offence severity classification).}
    \label{tab:singlemultiviewTableAction}
\end{table}

\begin{figure}[t]
  \centering
  \includegraphics[width=\linewidth]{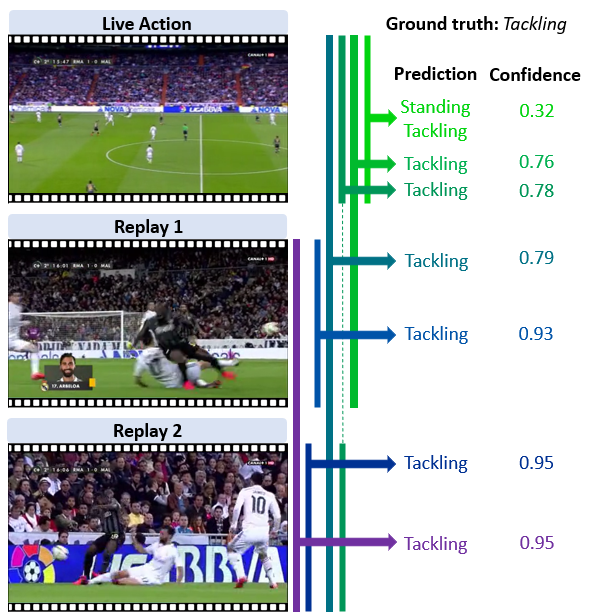}
\caption{
\textbf{Qualitative results.}
% We show a concrete example where our model's prediction is improved when increasing the number of views. The best performance is obtained thanks to the addition of the last view.
VARS predictions for different combinations of views as input.%by combining different number and types of views. 
The best performance is obtained with the two replay views.
}
\label{importance of the views}
\end{figure}

\subsection{Detailed analysis}

\mysection{Single \vs multi-view analysis.}
%In soccer, multiple views are often necessary to accurately determine the foul class, particularly for classes such as pushing, holding, and elbowing, which can be difficult to detect due to their subtle nature or obstruction from other players. 
%The use of multiple views, such as replays and live actions, increases the likelihood of detecting the correct foul class by providing different perspectives and information that may not be visible from a single view.
We now study the improvement of using multiple views over a single view.
To do so, we first created a subset of the test set for which we have clips with two replays and one live action. 
As evidenced by the top part of  
Table~\ref{tab:singlemultiviewTableAction}, the type of view has a significant impact on the VARS's ability to detect the correct type of foul. 
%In fact, replays tend to offer much more information compared to live action, as they are intended to provide the best possible angle for the spectator to understand the foul. 
%
Although the live-action view alone provides worse performance than the replays, combining the live-action view with a replay improves the accuracy slightly compared to using only the replay view in both tasks.
%However, in the case of balanced accuracy,it decreases when live action footage is added. 
This implies that even a poor-quality view can slightly improve the performance.
A highly informative view can boost the performance, as we can see by comparing the two replays with a single replay for the type of foul classification.
For the offence and severity classification, the VARS seems to benefit more from live actions compared to replays for the offense and severity classification task.
%Especially, in the case of the balanced accuracy where the replays lack behind the performance on the live actions.
One possible explanation is that for the live actions, the VARS takes into account the position of the action on the field, allowing it to learn that the likelihood of a 'No card' or 'Yellow card' is higher in specific areas of the field. 
%
%Regarding accuracy, the addition of one replay to the live action results in the best performance, while adding two replays decreases the model's overall performance.
%
%In general, the camera angle and positioning for replays are specifically chosen to provide the best possible perspective for spectators, making it easier for the model to identify the relevant features. 
%Figure \ref{dataset examples} illustrates the difference in information available between replays and live action.
%
For both tasks, we achieved better results by using multiple views, and for most of the metrics, the best performance was achieved by using a live-action clip with two replays.
This demonstrates the effectiveness of using multiple views to improve model performance in the type of foul and offence severity classification.

% \subsection{Qualitative results of VARS.}
In Figure~\ref{importance of the views}, we show the predictions of the foul classification models while changing the number and type of views.
By only using the live action, the VARS is not able to detect the correct type of foul, as confirmed in Table~\ref{tab:singlemultiviewTableAction}.
By adding $1$ or $2$ replays as input to the model, it is able to detect the foul class with a confidence score ranging from $76\%$ to $95\%$. 
By analyzing the confidence scores, we can see that the view has a big impact of the prediction, which agrees with the results found in Table \ref{tab:singlemultiviewTableAction}.

\mysection{Temporal analysis.}
We investigated the temporal context needed to identify fouls and offence severity.
% For the temporal analysis, we fixed the number of frames of the clips to 16 frames, where 8 frames are prior to the foul and 8 after the foul.
In particular, we increased the video length, by reducing the frame rate, in order to maintain the same number of frames to process.
Table \ref{tab:temporalAnalyse}, shows the results of the temporal analysis. 
We observed that as we increase the temporal context while decreasing the frame rates, the performance of our model decreases. 
This is likely because the most useful information for our classification tasks is concentrated within a narrow temporal window immediately preceding and following the foul. 
Adding more temporal context to the model results in the inclusion of frames that do not offer much additional information.
%In future work, we will analyze the impact on enlarging the temporal context while keeping the same frame rate.
%Unfortunately, the MViT feature extractor is at the moment only available with exactly 16 frames per clip.
% 
By default, we used a frame rate of 16 frames per second, with a temporal context of 1 second, which seemed to strike the best balance between capturing sufficient temporal information and excluding unnecessary frames.

\begin{table}[t]
    \centering
   \resizebox{1.0\linewidth}{!}{% <------ Don't forget this %
    \begin{tabular}{l|cccc}
    \toprule
% Temporal context& \textbf{3.2s}  & \textbf{2s}  &\textbf{1.33s} &\textbf{1s} \\
\textbf{Frame rate (FPS)}& \textbf{5}  & \textbf{8}  &\textbf{12} &\textbf{16} \\
\textbf{Temporal context} & \textbf{3.2s}  & \textbf{2.0s}  &\textbf{1.3s} &\textbf{1.0s} \\
% $T_{context}$& $[-1.6s;1.6s]$  & $[-1.0s;1.0s]$ &$[-0.67s;0.67s]$ &$[-0.5s;0.5s]$  \\
\midrule
Accuracy (Foul class.) &$0{.}36$ &$0{.}38$ &$0{.}44$ &$0{.}47$  \\
Accuracy (Off. sev. class.) &$0{.}39$ &$0{.}41$ &$0{.}43$  &$0{.}43$  \\
\bottomrule
    \end{tabular}
   }
    \caption{\textbf{Temporal analysis.}
    We experiment with various temporal context while maintaining a fixed number of 16 frames. 
    In all scenarios, we include 8 frames before and after the foul.}
    \label{tab:temporalAnalyse}
\end{table}

%Finally, we provide the performance of our multi-task VARS on different view configurations in Table~ \ref{tab:singlemultiviewTableOffenceSeverity}.
%Surprisingly, for the offence+severity classification, the live action helps the model much more compared to the case of the foul classification.
%Especially, in the case of the balanced accuracy where the replays lack behind the performance on the live actions.
%One possible explanation is that for the live actions, the VARS could take into account the position of the action on the field, allowing it to learn that the likelihood of a no card or yellow card is higher in specific areas. 
%Regarding accuracy, the addition of one replay to the live action results in the best performance, while adding two replays decreases the model's overall performance.
%
%For the balanced accuracy, the best performance is achieved using all three views.

\mysection{Per class analysis.}
We further analyze the performance per class.
The confusion matrices for both tasks are in the supplementary material.
We saw that performance varies considerably across classes.
For the fine-grained foul classification, the VARS struggles to distinguish between illegal arm movements due to their shared characteristics. It performs well in detecting ``Tackling'', but often confuses it with ``Dive'' due to the challenge of distinguishing genuine from deceptive actions in soccer games. The most difficult class for the VARS is ``Challenge'', as it shares visual similarities with many other classes, making proper generalization during training difficult.
Regarding offence classification, the VARS tends to make bad predictions in neighboring classes of the ground truth. For instance, it may classify a foul as ``Offence + Yellow card'' instead of ``Offence + No card''. However, the model struggles with ``Offence + Red card" due to the limited number of samples in the dataset.

\section{Conclusion}
\label{sec:conclusion}

In summary, our Video Assistant Referee System (VARS) has the potential to bring about a significant improvement in soccer refereeing by ensuring fairness and accuracy at all levels of professional and amateur play. VARS utilizes the latest advances in multi-view video analysis and provides referees with real-time feedback and assists them in making informed decisions that can impact the outcome of soccer games. To prove the effectiveness of VARS, we introduced a novel dataset, SoccerNet-MVFoul, that curates relevant fouls in soccer broadcasts from multiple views and includes foul properties. Our benchmarking results demonstrate that VARS can recognize foul characteristics based on multi-view video processing. 
By integrating the specific requirements of referees, VARS  offers an unbiased and reliable decision-making process for soccer matches. 
%We hope that our proposed system will encourage further research and development in automated soccer decision-making systems.

{
% \tiny
% \scriptsize
% \footnotesize
% \small
% \normalsize

\mysection{Acknowledgement.}
This work was partly supported by the King Abdullah University of Science and Technology (KAUST) Office of Sponsored Research through the Visual Computing Center (VCC) funding and the SDAIA-KAUST Center of Excellence in Data Science and Artificial Intelligence (SDAIA-KAUST AI). 
A. Cioppa is funded by the F.R.S.-FNRS.

}

\cleardoublepage

%%%%%%%%% REFERENCES
{\small
\bibliographystyle{ieee_fullname}
\bibliography{bib/abbreviation-short,bib/activity,bib/dataset,bib/labo,bib/put-new-refs-here,bib/soccer,bib/sports,bib/bib}
}

\cleardoublepage

\section{Supplementary}

\subsection{Video assistant referee system software}\label{label:vars interface}

The design of both the VARS annotator and the VARS interface draws inspiration from the VAR room.
To enhance the user experience, a grid layout is used to display all available perspectives synchronously.
The objective is to have an easy to use interface to annotate and to predict different properties of an action.

\begin{figure*}[htbp]
  \centering
  \begin{subfigure}{\textwidth}
  \includegraphics[width=\textwidth]{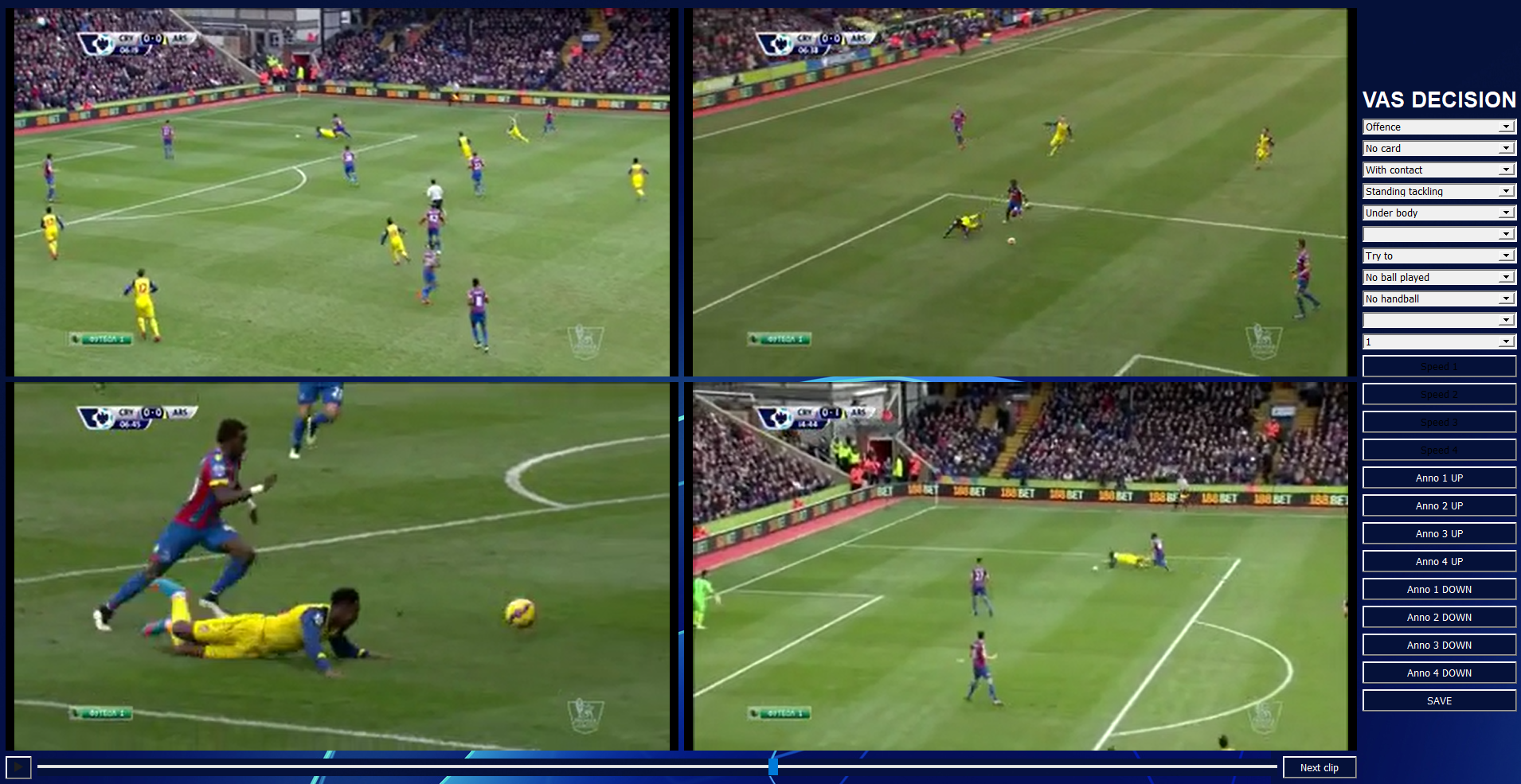}
    \caption{\textbf{VARS annotator.} The annotator may browse simultaneously the synchronised videos either at regular speed or frame by frame. He can annotate all 10 properties and adjust the annotated point of contact for each clip separately and temporal align the different clips by modifying the speed
            and offset of the clips.}
    \label{vas_interface}
\end{subfigure}
\hfill
\begin{subfigure}{\textwidth}
    \includegraphics[width=\textwidth]{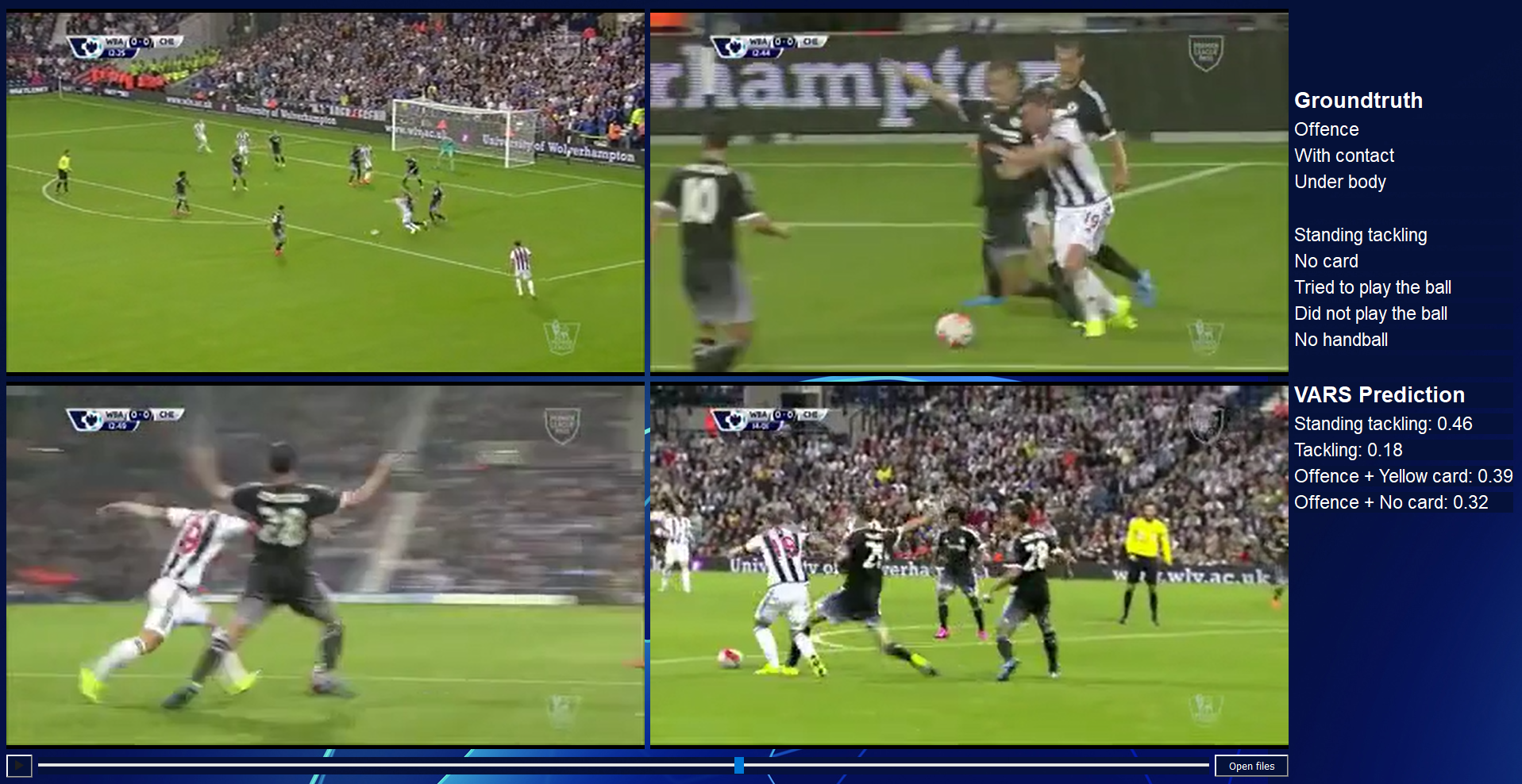}
   \caption{\textbf{Video Assistant Referee System interface.}
    The interface of the VARS shows the ground truth of the action and his top 2 predictions for the foul classification task, and the offence and severity classification task. In this example, the VARS correctly predicts the foul resulting in a penalty. }
    \label{fig:VARS}
\end{subfigure}
            \caption{\textbf{Views of the VARS interfaces.} (a) shows the interface for the annotation process.
            (b) shows the interface of the VARS for displaying its results.
            \label{interfaces}}
\end{figure*}

\mysection{VARS annotator.}
To increase the speed of the annotation task, we build a VARS annotator (Figure~\ref{vas_interface}), which shows all the available clips of an action simultaneously. 
The VARS annotator allows for individual adjustment of the annotated moment for each clip to achieve temporal alignment, speed adjustment for replays, and annotation of all the properties.

\mysection{VARS interface.}
The VARS interface has the same interface as the annotator. 
 It enables easy access to all available perspectives for a particular action.
 The multi-task VARS, which achieved the best results on the test set, is built directly into the interface, allowing for immediate analysis of selected videos.
The VARS interface offers top two predictions for the type of foul classification, as well as the offense and severity classification for the selected videos. 
Furthermore, for each prediction, the VARS interface shows the confidence score of his prediction.

In Figure~\ref{fig:VARS}, we can see an example of how the VARS interface looks.
In this example, the VARS correctly predicts the type of foul, and tells  that the action was indeed a foul, which leads in a penalty for the attacking team in this example.
For the severity classification, the VARS was uncertain whether to assign no card or a yellow card.

\subsection{Dataset}

\subsubsection{Property explanations} \label{supp_dataset}
\begin{figure*}
\centering
\begin{subfigure}{\textwidth}
    \includegraphics[width=\textwidth]{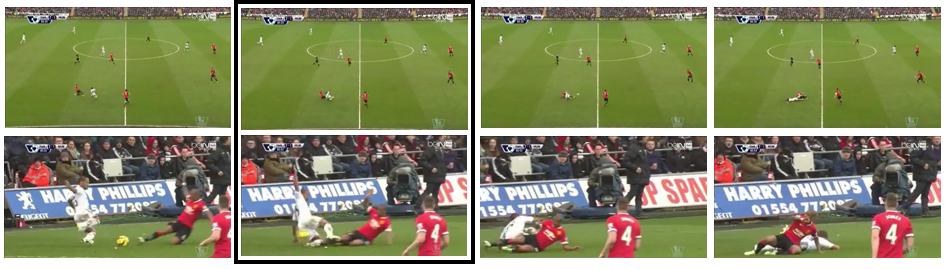}
    \caption{``Offence'', ``Tackling'', ``Yellow card'', ``With contact'', ``Under body'', ``/'', ``Played the ball'', ``Tried to play the ball'', ``No handball''}
    \label{fig:illegalUseOfArms_Supplementary}
\end{subfigure}
\hfill
\begin{subfigure}{\textwidth}
  \includegraphics[width=\textwidth]{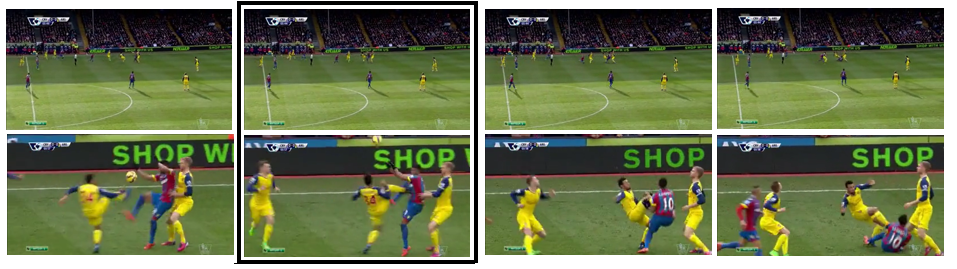}
    \caption{``Offence'', ''High leg'', ''Red card'', ``With contact'', ``Under body'', ''/'', ``Ball is not played'', ``Tried to play the ball'', ``No handball'' and ``No handball offence'' }
    \label{fig:tackling_Supplementary}
\end{subfigure}
\hfill
\begin{subfigure}{\textwidth}
    \includegraphics[width=\textwidth]{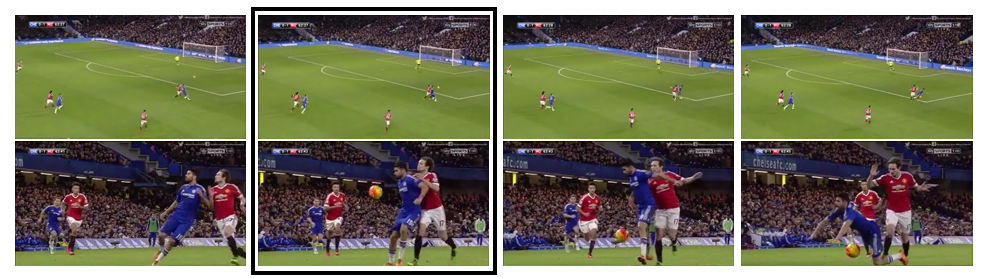}
   \caption{``Offence'', ``Challenge'', ``No card'', ``With contact'', ``Upper body'', ``Use of shoulder'', ``Ball is not played'', ``Tried to play the ball'', ``No handball'' and ``No handball offence'' }
    \label{fig:no card_Supplementary}
\end{subfigure}
\hfill
\begin{subfigure}{\textwidth}
    \includegraphics[width=\textwidth]{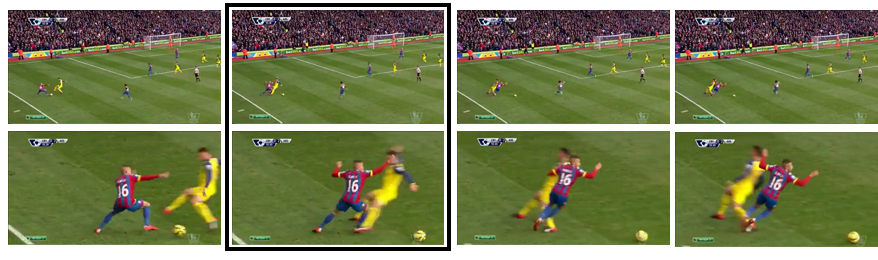}
    \caption{``Offence'', ``standing tackling'', ``Yellow card'', ``With contact'', ``Under body'', ''/'', ``Did not play the ball'', ``Tried to play the ball'', ``No handball'' }
    \label{fig:red card_Supplementary}
\end{subfigure}
        
\caption{\textbf{Dataset overview and ground truth.} We annotated the exact frame where the point of contact happens (depicted by the back box).
           }
 \label{fig:dataset examples supp}
\end{figure*}

This section explains in detail each property. Furthermore, we illustrate multiple examples of our dataset in Figure~\ref{fig:dataset examples supp}.

\mysection{Was it a foul?}
    The task of identifying fouls is a critical and challenging aspect of the role of a referee and VAR. They must determine whether an action is a foul or not, in accordance to the laws of the game ~\cite{IFAB2022Laws}. For each action, we determined whether an action is an (i) offence (an action which breaks/violates the Laws of the Game ~\cite{IFAB2022Laws}), (ii) no offence (did not break/violate the Laws of the Game), or (iii) between (if the action lays inside a grey area). In some cases, both decisions may be correct and the final decision depends on the interpretation of the rules of an individual.
    
    It is worth mentioning that, for each clip of the same foul, we make the same annotation. During the annotation process, we looked at all the clips and took a global and final decision which is the same for each clip of the same foul. 

\mysection{Was there any contact?}
    Another important property which we annotated was if there was any contact between two players during an action. We annotated for each foul (i) with contact (if there was contact between players), or (ii) without contact (if there was no contact).
    This property is important because a foul with contact such as a tackling, holding, elbowing result in a direct free kick, while a foul without contact such as a simulation or dangerous play will result in an indirect free kick. 

\mysection{Did the player touch the ball with his hand/arm?}
    This property annotates if the player touches the ball, deliberately or not, with his hand or arm. We annotate (i) handball (if a players touches the ball with his hand/arm), or (ii) no handball (if the ball did not touch the hand/arm). This property only states whether the ball touched the hand or arm and does not indicate whether the handball is punishable or not.
    An important note to make is that the upper boundary of the arm is in line with the bottom of the armpit~\cite{IFAB2022Laws}.

\mysection{Was the upper or under body used in the action?} 
    This property annotates which part of the body was used during an action. We differentiated between (i) under body (which corresponds to the use of the food or the leg), or (ii) upper body (which corresponds to the use of shoulder or the use of arms).

\textbf{With which part of the upper body was the action made?} 
    In the case where we annotated the previous property with "upper body", we further split between (i) use of shoulders, or (ii) use of arms.

\mysection{Class of the action.} 
    This property annotates the type of action. In total, we have 9 different classes:
\begin{enumerate}
\item 
\textbf{Tackling.} The sliding movement of a player towards an opponent who is in possession of the ball and legally or illegally using his foot or leg to try to take the ball away. 
\item 
\textbf{Standing tackling.} The movement (not sliding) of a player towards an opponent who is in possession of the ball and legally or illegally using his foot or leg to try to take the ball away. 
\item 
\textbf{Holding.} Occurs when a player’s contact with an opponent’s body or equipment impedes the opponent’s movement~\cite{IFAB2022Laws}.
\item 
\textbf{Pushing.} The action of using the upper body to push an opponent away.
\item 
\textbf{Challenge.} Physical challenge against an opponent, using the shoulder and/or the upper arm~\cite{IFAB2022Laws}.
\item 
\textbf{Elbowing.} The use of arms (and frequently the elbows) as a tool or a weapon to gain an unfair advantage in aerial challenges, physical battles, to create space or to intimidate other players.
\item     
\textbf{High leg.} A movement where a player swings his foot close to and above the waist of an opponent.
\item     
\textbf{Dive.} An action which creates a wrong/false impression that something has occurred when it has not, committed by a player to gain an unfair advantage. ~\cite{IFAB2022Laws}
\item   
\textbf{Don't know.} Corresponds to anything which can not be classified in one of the classes above.
     \end{enumerate}

\mysection{How severe was the foul?} 
    For each foul, we annotated the severity of the foul by a scale from 1 to 5: 
\begin{itemize}
\item \textbf{1:}  
a careless foul which is when a player shows a lack of attention or consideration when making a challenge or acts without precaution. No disciplinary sanction is needed. (No card)~\cite{IFAB2022Laws} 
\item \textbf{2:} 
a borderline foul between careless and reckless. We are in a grey area where both ``no card'' or ``yellow card'' would be correct.
\item \textbf{3:} 
a reckless foul which is when a player acts with disregard to the danger to, or consequences for, an opponent and must be cautioned. (Yellow card)~\cite{IFAB2022Laws} 
\item \textbf{4:}  
a borderline foul between reckless and violent. We are in a grey area where both ``yellow card'' or ``red card'' would be correct.
\item \textbf{5:}  
a violent foul where a player exceeds the necessary use of force and/or endangers the safety of an opponent and must be sent off. (Red card) ~\cite{IFAB2022Laws}.
\end{itemize}

\mysection{Did the player try to play the ball?}
    When a player commits an offence against an opponent within their own
    penalty area which denies an opponent an obvious goal-scoring opportunity
    and the referee awards a penalty kick, the offender is cautioned if the offence
    was an attempt to play the ball; the offender is sent off if there was no possibility to play the ball~\cite{IFAB2022Laws}.
    We annotated, (i) ``Yes'', if the player tried to play the ball, or ``No'', if there was no possibility to play the ball.

\mysection{Did the player play the ball?}
    The final property annotates (i) ``Yes`' if the defender touches the ball, (ii) ``No'' when the defender did not play the ball, or (iii) ``Maybe'' in the case where the quality of the video or the viewpoint on the foul is not sufficient to determine if the player touched the ball or not.

% Action classification: Performance per class

\begin{table*}[t]
    \centering
    \resizebox{\linewidth}{!}{% <------ Don't forget this %
    \begin{tabular}{lc|lc|lc|lc|lc}
    \toprule
        \multicolumn{2}{c|}{\bf Fouls} &  \multicolumn{2}{c|}{\bf Severity} &  \multicolumn{2}{c|}{\bf Offence} &  \multicolumn{2}{c|}{\bf Handball} &  \multicolumn{2}{c} {\bf Handball offence} \\ \midrule
       \bf Class & \bf Prob. & \bf Class & \bf Prob. & \bf  Class & \bf Prob. & \bf  Class & \bf Prob. & \bf  Class & \bf Prob.\\ \midrule
        St. tackling & 0.43 &No card  &0.55  &Offence  &0.85 &Yes &0.99 &Yes &0.82  \\ 
        Tackling     & 0.15 &Yellow card &0.26  &No offence  &0.10 &No &0.01 &No &0.18 \\
        Challenge     & 0.13 &NC/YC  &0.15  &Between  &0.03 & & & & \\ 
        Holding     & 0.12  & YC/RD  &0.02 &  & & & & & \\
        Elbowing     & 0.05  &Red card  &0.01  &  & & & & & \\
        High leg     & 0.03  &  &  &  & & & & & \\
        Pushing     & 0.02 &  &  &  & & &\\
        Dive     & 0.01  &  &  &  & & & & &\\
    \bottomrule
    \end{tabular}
    
    }
    \caption{Distribution of classes in our SoccerNet-MVFouls dataset.}
    \label{tab:distribution_all1}
\end{table*}

% Action classification: Performance per class

\begin{table*}[t]
    \centering
    \resizebox{\linewidth}{!}{% <------ Don't forget this %
    \begin{tabular}{lc|lc|lc|lc|lc}
    \toprule
        \multicolumn{2}{c|}{\bf Bodypart} &  \multicolumn{2}{c|}{\bf Upperbody part} &  \multicolumn{2}{c|}{\bf Try to play the ball} &  \multicolumn{2}{c|}{\bf Played the ball} &  \multicolumn{2}{c} {\bf Contact} \\ \midrule
       \bf Class & \bf Prob. & \bf Class & \bf Prob. & \bf  Class & \bf Prob. & \bf  Class & \bf Prob. & \bf  Class & \bf Prob.\\ \midrule
        Upperbody & 0.36 &Arms  &0.66  &Yes  &0.92 &Yes &0.10 &With &0.97 \\ 
        Underbody    & 0.64 &Shoulder &0.33  &No  &0.08 &No &0.87 &Withou &0.03 \\
             &  &  &  &  & &Maybe &0.02 & & \\ 

    \bottomrule
    \end{tabular}
    }
    \caption{Distribution of classes in our SoccerNet-MVFouls dataset.}
    \label{tab:distribution_all2}
\end{table*}

\subsubsection{Dataset distribution}
The distribution of all classes is provided in Tables ~\ref{tab:distribution_all1} and ~\ref{tab:distribution_all2}.
Most of the properties in the dataset have a high degree of imbalance, particularly the ``Handball'' property that determines if a player has made contact with the ball using their arm or hand. 
Nearly 99\% of the actions recorded in the dataset do not involve any handball. Similar imbalances are observed in the ``Contact'', ``Try to play the ball'', ``Played the ball'', and ``Offence'' properties.

The ``Bodypart'' and ``Upperbody part'' properties are relatively less unbalanced, with a distribution of approximately 66\% for the superior class and 34\% for the inferior class.

\section{Experiments} \label{supplementary experiments}

\subsection{Per class analysis}

\mysection{Foul classification task}

\begin{figure}[htbp]
   \centering
   \includegraphics[width=0.8\linewidth]{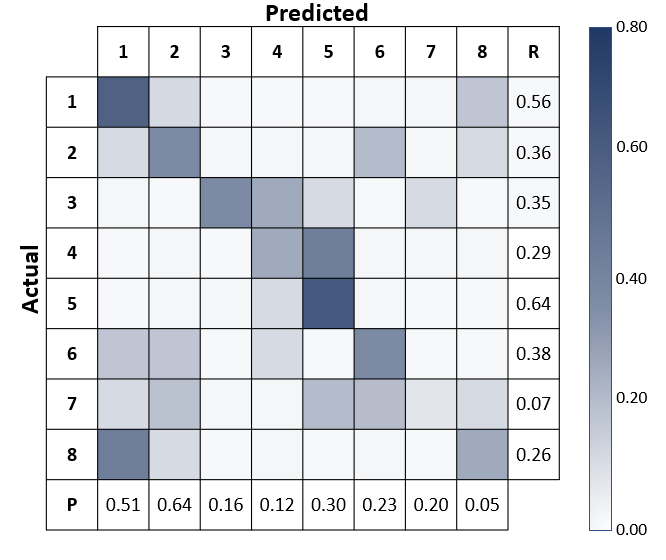}
             \caption{\textbf{Confusion matrix for the type of foul classification.} 
            The VARS demonstrates good performance in classifying ``Standing Tackling'', ``Tackling'', and ``Elbowing''. However, the model struggles with ``Challenge'' and frequently confuses it with other classes.
             $1$: Standing Tackling, $2$: Tackling, $3$: High Leg, $4$: Pushing, $5$: Elbowing, $6$: Holding, $7$: Challenge, $8$: Dive, R: Recall and P: Precision.} 
             \label{confusion_matrix_action} 
 \end{figure}

  The performances for each class are summarized in the confusion matrix shown in   Figure~\ref{confusion_matrix_action}. %, while the confusion matrix is provided in the supplementary materials (see section \ref{confusion matrix action}).
 Our analysis shows that the performance varies considerably across classes.
 The VARS often confuses all the illegal use of arm classes, like ``Holding'', ``Pushing'', and ``Elbowing'' as these fouls share some common characteristics and can involve similar physical movements. 
 The model performs well in detecting ``Tackling'', but confuses it often with ``Dive'' as it struggles to distinguish between genuine and deceptive actions, which can be challenging due to the complex and dynamic nature of soccer games.
 However, the most challenging class for the VARS is ``Challenge'', which shares visual similarities with many other classes, making it difficult for the system to generalize properly during training.

%In contrast to a ``Tackling'' or ``Standing Tackling'', a ``Dive'' refers to a player's intentional act of simulating contact with an opponent to deceive the referee into awarding a foul or penalty. 
%This deceptive behavior poses a significant challenge for our VARS, as it requires the system to accurately identify and interpret the subtle body movements and intentions of the player. 
%our method struggles to detect this foul type reliably, as the detection of a ``Dive'' heavily relies on the system's ability to distinguish between genuine and deceptive actions, which can be challenging due to the complex and dynamic nature of soccer games.
%Classifying between ``Holding'', ``Pushing'', and ``Elbowing'' in soccer is challenging for the VARS as these fouls share some common characteristics and can involve similar physical movements.

\mysection{Offence and severity classification task}

\begin{figure}[htbp]
   \centering
   \includegraphics[width=0.7\linewidth]{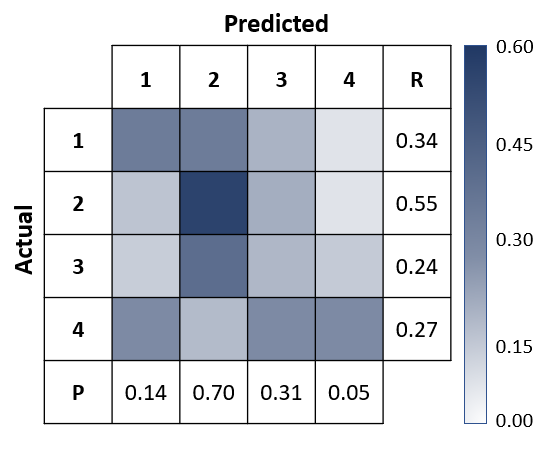}
             \caption{\textbf{Confusion matrix for the offence and severity classification.} The VARS shows good performance for the ``Offence + No Card'' class. The model confuses the classes ``No Offence'' and ``Offence + Yellow Card''  with ``Offence + No Card''. 
             For the class ``Offence + Red Card'' the model is not able to provide good results due to the low amount of samples in the dataset.
             1: No offence, 2: Offence + No card, 3: Offence + Yellow card, 4: Offence + Red card, R: Recall and P: Precision.} 
             \label{confusion_matrix_offence_main}
 \end{figure}

Figure ~\ref{confusion_matrix_offence_main} displays the confusion matrix for the offence and severity classification, revealing that the model frequently confuses classes with their neighboring classes. 
For instance, when the ground truth is ``Offence + No card'', the VARS often mistakes it for ``No offence'' or ``Offence + Yellow card''.

Indeed, the visual similarity between all the classes, especially with the neighbor classes, is very high.
Small details, which very often can only be seen in a couple of frames, can differ between the actual class.
Factors such as the speed of the foul, the point of contact, or the intention of playing the ball are critical criteria for deciding which class an action corresponds to.
However, these criteria can be challenging to spot for a model and to differentiate between the different classes.
Furthermore, there is only a small number of instances of ``Offence + Red card'' in the dataset, making it more challenging for the model to generalize.
Despite all these difficulties, the VARS is still able to achieve an accuracy of $0.43$.

\end{document}